\documentclass[10pt,twocolumn,letterpaper]{article}

\usepackage{cvpr}       
\usepackage{listings}
\usepackage{xcolor}
\usepackage{enumitem}
\usepackage{array}
\usepackage{amsmath}
\usepackage{amssymb}
\usepackage{booktabs}
\usepackage{dsfont}
\usepackage{graphicx}
\usepackage{makecell}
\usepackage{multirow}
\usepackage{pifont}
\usepackage[group-separator={,}]{siunitx}
\usepackage{tabularx}
\usepackage{xcolor}
\usepackage{caption}
\usepackage{soul}
\usepackage{changepage,threeparttable}
\usepackage{colortbl}
\usepackage{algpseudocode}
\usepackage{algorithm}
\usepackage{xspace}

\usepackage{mathtools}
\usepackage{xfrac}
\usepackage{url}

\usepackage{catchfile}
\usepackage{longtable}
\usepackage{tabu}
\usepackage{supertabular}
\usepackage{multicol}
\usepackage{enumitem}

\newcolumntype{Y}{>{\centering\arraybackslash}X}
\usepackage[accsupp]{axessibility}  

\definecolor{color_1}{RGB}{255,0,128}
\definecolor{color_2}{RGB}{0,128,128}
\definecolor{color_3}{RGB}{0,128,0}
\definecolor{color_4}{RGB}{128,0,0}
\definecolor{color_5}{RGB}{128,0,128}
\definecolor{cadetgrey}{RGB}{0.57, 0.64, 0.69}

\usepackage{pgfplots,pgfplotstable}
\usepgfplotslibrary{groupplots}
\pgfplotsset{compat=1.18}
\newif\ifarxiv
\arxivtrue

\definecolor{cvprblue}{rgb}{0.21,0.49,0.74}
\definecolor{red}{rgb}{0.90,0.0,0.20}
\definecolor{green}{rgb}{0.49,0.76,0.00}
\definecolor{rulecolor}{HTML}{888888}
\definecolor{first}{HTML}{a9c98f}
\definecolor{second}{HTML}{eaf2e3}

\definecolor{third}{HTML}{eaf2e3}
\definecolor{beige}{HTML}{f2f0e9}
\definecolor{sunny}{HTML}{fffce8}

\newcommand{\method}{HOIWorld\xspace}
 
\newcommand{\cI}{\cellcolor{first}}
\newcommand{\cII}{\cellcolor{second}}

\newcommand{\ours}{\cellcolor{beige}}

\usepackage[pagebackref,breaklinks,colorlinks,allcolors=cvprblue]{hyperref}
\usepackage{amsmath}
\usepackage{colortbl}
\usepackage{makecell}

\arrayrulecolor{rulecolor}
\usepackage[T1]{fontenc} 
\usepackage{iftex}
\ifLuaTeX
\protected\def\pdfmapline {\pdfextension mapline }
\fi
\pdfmapline{+goodlife < Goodlife.ttf <T1-WGL4.enc}

\title{Peeled Depth as a Unified Action Interface for Dexterous World Models}

\author{Alara Dirik$^{1}$, Rolandos Alexandros Potamias$^{1}$, Stefanos Zafeiriou$^{1}$
\\ $^1$Imperial College London \hspace{5em} 
}

\begin{document}
\twocolumn[{%
\renewcommand\twocolumn[1][]{#1}%
\maketitle
\begin{center}
    \centering
    \captionsetup{type=figure}
    \vspace{-1.5em}
    \includegraphics[width=\textwidth]{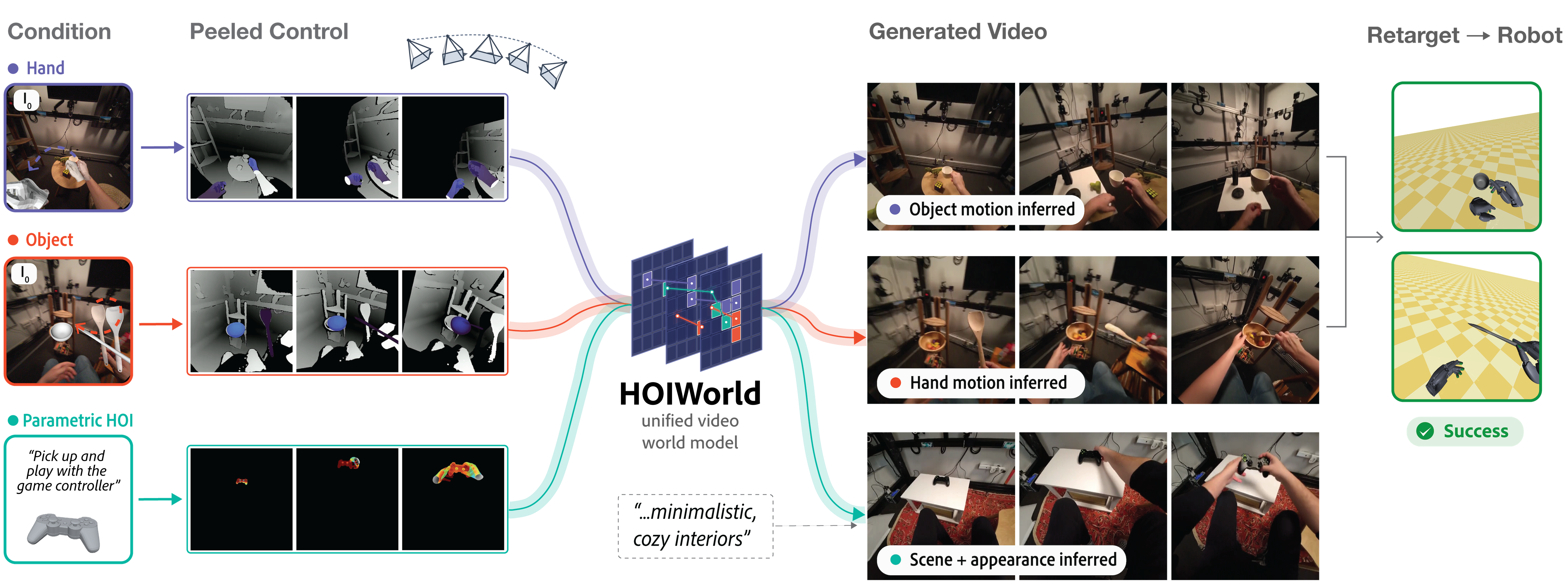}
    \vspace{-1.5em}
    \captionof{figure}{We propose \method, a unified dexterous world model that generates egocentric hand-object interaction videos from a camera trajectory and a bimanual hand trajectory, a 6-DoF object trajectory, or text-generated parametric HOI, each rendered as a peeled control video. The generated rollouts infer the unspecified object or hand motion, and can be successfully retargeted to a dexterous robot hand.}
\end{center}%
}]

\begin{abstract}
Emerging dexterous world models predict how a scene evolves under hand-object interaction, but are driven by hand trajectories alone. Hand trajectories offer fine-grained control, but the natural command is often what should happen to the object. We present \method, a flexible dexterous world model that can be driven by bimanual hand trajectories, 6-DoF object trajectories, or both, and imagines whatever the action leaves unspecified. Our key idea is a geometric action compiler that renders scene, camera, and action into a hybrid control video, in which the controlled hands or objects appear as \emph{peeled depth} that retains surfaces hidden by occlusion, such as self-occluded fingers and grasped objects, and drives our action-biased attention. Trained only on HOT3D, \method outperforms specialized baselines in both modes on HOT3D and, zero-shot, on EgoDex and OakInk2. Under hand control, it achieves the lowest hand pose error and reduces camera rotation error by 67\% over the strongest dexterous world model; under object control, it reduces object pose error by 89\% over the strongest object-conditioned generator, while enabling 4D sim-to-real generation and robot retargeting. 
\end{abstract}
\section{Introduction}
\label{sec:intro}

We interact with the world primarily through our hands, where grasping, lifting, using and moving objects change the state of our surroundings and determine our future observations. A model that predicts how a scene evolves under dexterous interaction is therefore valuable both for modeling the physical world and for transferring that knowledge to robots. Video generative models are increasingly used to instantiate such world models: given an observation and an action, they predict how the world would look if that action were taken~\cite{Ha2018WorldM, Yang2023LearningIR, Bruce2024GenieGI, Agarwal2025CosmosWF}. In dexterous manipulation, such a model lets an agent ask counterfactual questions before acting (e.g. \textit{what happens to this scene if the hand closes around the mug, or if the mug is lifted and tilted toward the sink?}). We argue that dexterous world models are natural successors to general purpose, egocentric hand-object interaction (HOI) video generators, which are already used as planners and simulators for robot learning and action retargeting~\cite{Du2023LearningUP, jang2025dreamgen, Sharma2026WorldGymnastTR, Zhao2025TASTERobAV}, as well as text-to-4D HOI methods that generate interaction sequences of parametric hand meshes with a limited set of CAD objects~\cite{Cha2024Text2HOIT3, Christen2024DiffH2ODS, Pratikaki2026Dex2HOIDB}. Recent dexterous world models have made notable progress toward this goal, generating photorealistic HOI videos from a reference frame, the camera trajectory, and the per-frame 3D poses of both hands~\cite{Li2026EgocentricWM,Hao2026EgoSimEW,Xie2026GeneratedRH,Chen2026HandsOnWorldUE}.

However, existing dexterous world models largely fix the action space to the hands. Hand trajectories provide fine-grained control over how an interaction is performed, but they are costly to specify: they require all joints at every frame and are tied to a particular parametric hand model. In many planning settings, the more natural command is instead what should happen to the object, and a dexterous world model should accept both. Object trajectories are compact, easy to edit, largely embodiment-agnostic, and increasingly predicted by dedicated models from images, goals, or language~\cite{Soraki2026ObjectForesightPF,Saroha2026EgoFlowGF,Zeng2026GMTGM}. Object-trajectory conditioned video generators accept such commands~\cite{Fu20243DTrajMasterM3,Chu2025WanMoveMV}, but are largely limited to exocentric settings, mostly driven by 2D tracks, and prone to failure under the heavy occlusion common in hand-object manipulation. 
Hand- and object-driven generation have therefore developed as separate lines of work, and existing hand-driven models~\cite{Li2026EgocentricWM,Hao2026EgoSimEW,Xie2026GeneratedRH,Chen2026HandsOnWorldUE} feed the hand trajectories, camera trajectory, and scene as separate signals through dedicated encoders, forcing the model to infer how they relate in 3D.

This raises a natural question: can a single dexterous world model be driven by either the bimanual hand trajectories or the object trajectory and imagine everything left unspecified, namely how the object responds to a given hand trajectory, or which hand motion and contacts yield a given object trajectory? We observe that these heterogeneous actions become the same kind of signal once instantiated geometrically: a hand pose defines a posed hand surface, an object pose a posed rigid surface, and a camera trajectory the viewpoint from which both, together with the scene, are observed. Rendering them into one view-aligned control video places scene and action in a shared metric 3D frame, lets a pretrained depth-controlled video model read them without action-specific encoders, and expresses the control mode purely by which geometry is rendered. Naively rendered depth, however, is inadequate: it stores only the nearest surface along each camera ray, so self-occluded fingers and grasped objects vanish and distinct hand or object configurations collapse to nearly identical depth. Moreover, the control signal must be constructible from a single reference image at deployment, since scene geometry reconstructed from future frames would act as privileged conditioning.

Building on these observations, we propose \method, a dexterous world model controlled through a hybrid geometric representation that we call peeled action geometry. From the reference frame alone, we lift the visible scene to 3D and re-render it as depth along the target camera trajectory, leaving holes where later views reveal unseen regions, so that camera motion enters as apparent scene motion and the model is shown exactly where new content must be imagined. Our key insight is to render the controlled hands or objects on top of this scene as \emph{peeled depth}: multi-hit raycasting retains every intersection of a camera ray with the action meshes, so fingers hidden by self-occlusion and object surfaces hidden by the grasp remain in the control signal. We encode these hidden surfaces in the color channels that a grayscale depth control leaves unused, so the hybrid signal enters a pretrained depth-control branch unchanged. Our only architectural change is an \emph{action-biased attention}: the same raycast yields a mask of the controlled geometry, and tokens the model must complete attend more strongly to it, coupling the inferred hands, objects, and disoccluded scene to the action. A single model, fine-tuned from Wan2.2-Fun-Control~\cite{Wang2025WanOA} on HOT3D~\cite{Banerjee2024HOT3DHA}, supports hand-, object-, and joint hand-object control without task identifiers or action-specific encoders.

\begin{figure*}[t]
\centering
\includegraphics[width=\textwidth]{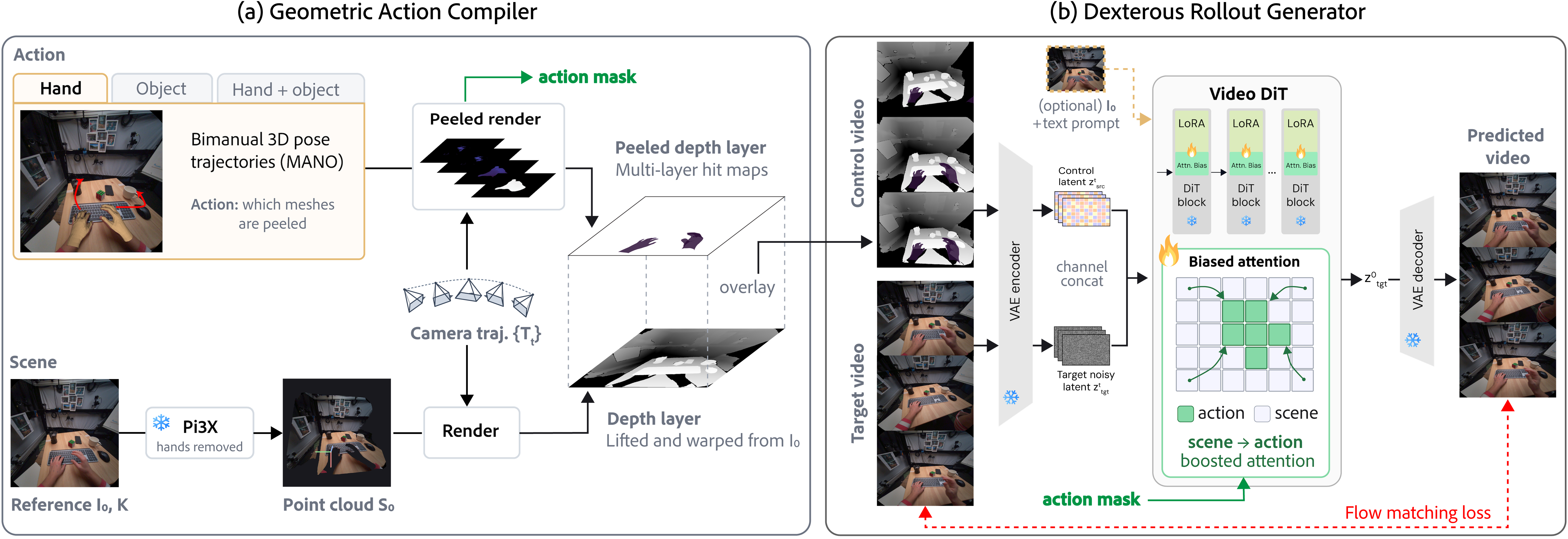}
\caption{\textbf{Overview of \method.} (a) A geometric action compiler converts heterogeneous action sources into one control video. From the reference frame $\mathbf{I}_0$, scene geometry is reconstructed once and re-rendered along the target camera trajectory; controlled hand and/or object meshes are overlaid as peeled depth so that both visible and self-occluded action geometry are retained. Hand-only, object-only, and joint hand--object control differ only in the geometry routed to the peel. (b) The resulting control video is channel-concatenated with the target video as input to a pretrained video DiT in every setting for finetuning with a learnable action-biased attention.}
\vspace{-1.25em}
\label{fig:pipeline}
\end{figure*}

Trained only on HOT3D, \method outperforms dedicated hand-conditioned world models on HOT3D and, zero-shot, on EgoDex~\cite{Hoque2025EgoDexLD}, improving PSNR from 15.33 to 19.01~dB and reducing camera rotation error from $81.6^\circ$ to $26.6^\circ$ over the strongest of them on HOT3D. Under object control, the same model outperforms object-trajectory video generators on HOT3D and, zero-shot, on OakInk2~\cite{Zhan2024OakInk2A}, reducing ADD-S from 25.00 to 2.66~cm relative to Wan-Move. Its rollouts can be executed by an off-the-shelf dexterous retargeting pipeline~\cite{Paliwal2026DoAI} with success rates within six points of real video, and under joint control it turns hand-object interactions generated from text~\cite{Pratikaki2026Dex2HOIDB} into realistic video without a reference image. Our main contributions are:
\begin{itemize}[leftmargin=1.2em,itemsep=1pt, topsep=2pt]
\item \method, to the best of our knowledge, the first dexterous world model driven by bimanual hand trajectories, 6-DoF object trajectories, or both within a single model.
\item Peeled action geometry, a deployable, occlusion-preserving control representation, and action-biased attention derived from the same raycast.
\item State-of-the-art hand- and object-conditioned generation with zero-shot transfer, and downstream use for dexterous retargeting and sim-to-real HOI video.
\end{itemize}
\vspace{-0.5em}
\section{Related Work}
\label{sec:related}

\paragraph{World models for embodied interaction.} A world model predicts how an environment evolves under an action. Video generative models are well suited to this role: large-scale pretraining equips them with strong physical priors, such as object permanence, articulated motion, and contact, and they represent state directly in pixels. Recent work conditions video models on game controls and camera motion~\cite{Bruce2024GenieGI, Agarwal2025CosmosWF}, robot actions~\cite{Yang2023LearningIR, Agarwal2025CosmosWF}, and egocentric full-body poses~\cite{Bai2025WholeBodyCE, Pallotta2025EgoControlCE}, and deploys them for robot learning, where they plan actions~\cite{Du2023LearningUP}, generate training data~\cite{Agarwal2025CosmosWF, jang2025dreamgen, Bharadhwaj2024Gen2ActHV}, and train policies with reinforcement learning~\cite{Sharma2026WorldGymnastTR}. Dexterous world models extend this conditioning to articulated hands, taking the camera trajectory and the per-frame poses of both hands, and differ mainly in how the hands are encoded: EgoHOI~\cite{Li2026EgocentricWM} and Hand2World~\cite{Wang2026Hand2WorldAE} rasterize the posed hand meshes, Generated Reality~\cite{Xie2026GeneratedRH} encodes joint angles, HandsOnWorld~\cite{Chen2026HandsOnWorldUE} encodes rays of the visible hand surface, and JointControl~\cite{Zhang2026ControllableEV} splats sparse 3D joints into heatmaps, down-weighting joints it estimates to be occluded; all observe the scene only in the first frame. DWM~\cite{Kim2025DexterousWM} and EgoSim~\cite{Hao2026EgoSimEW} also render the static scene along the camera trajectory, but still provide the hands separately, as mesh renders and 2D keypoints, respectively. HandWorld~\cite{Sun2026HandWorldHU} instead generates video and hand poses jointly, and DexWM~\cite{Goswami2025WorldMF} autoregressively predicts DINOv2 features rather than pixels. \method addresses three limitations these models share: it accepts object as well as hand trajectories, renders the controlled hands or objects and the scene into one control video in a shared 3D frame instead of separate signals, and peels the controlled geometry so that surfaces hidden by self-occlusion or the grasp, such as occluded fingers or object parts, are retained.

\paragraph{Hand-object interaction generation.} While world models simulate the consequences of a given action, HOI generation methods generate the interaction itself, typically as video from a reference image and a text or 2D signal: InterDyn~\cite{Akkerman2024InterDynCI} animates the image from binary hand masks, while TASTE-Rob~\cite{Zhao2025TASTERobAV} and SCAR~\cite{Yan2025OpenworldHI} start from a task description, the latter also predicting hand-object contours, contacts, and depth; none accepts 3D hand or object trajectories. HVG-3D~\cite{Chen2026HVG3DBR} comes closest by conditioning on hand and object point clouds, but encodes them as tokens that are not pixel-aligned with the video, requires a real reference image even for simulated input, and shows simulation-driven results only qualitatively. Conversely, 3D methods generate MANO hand sequences and 6-DoF object poses from text~\cite{Cha2024Text2HOIT3, Christen2024DiffH2ODS, Pratikaki2026Dex2HOIDB} or forecast 6-DoF object motion~\cite{Soraki2026ObjectForesightPF,Saroha2026EgoFlowGF,Zeng2026GMTGM}, but generate no video. \method connects the two: it renders such 3D motion into egocentric video, without a reference image under joint control, and its rollouts can be retargeted to robot hands~\cite{Paliwal2026DoAI} (\cref{sec:exp:apps}).

\vspace{-1em}
\paragraph{Controllable video generation.} Video diffusion models~\cite{Wang2025WanOA} accept spatially aligned conditions such as depth through ControlNet-style~\cite{Zhang2023AddingCC} branches or channel concatenation~\cite{Wang2023VideoComposerCV}, as in our depth-controlled backbone~\cite{Wang2025WanOA,aigc_apps_VideoX_Fun_2026}. For object motion, 2D drags or tracks~\cite{Wang2023MotionCtrlAU,Zhang2024ToraTD} leave depth and out-of-plane rotation ambiguous; Wan-Move~\cite{Chu2025WanMoveMV} and LeviTor~\cite{Wang2024LeviTor3T} add depth cues but remain 2D-driven, Diffusion as Shader~\cite{Gu2025DiffusionAS} conditions on rendered 3D point tracks, 3DTrajMaster~\cite{Fu20243DTrajMasterM3} injects 6-DoF entity poses into text-to-video generation, and LooseControlVideo~\cite{Bhat2026LooseControlVideoDV} renders oriented 3D boxes with depth-ordered occlusion, while camera motion is typically encoded via Pl\"ucker embeddings~\cite{He2024CameraCtrlEC} or by re-rendering a point cloud lifted from the input frame~\cite{Ren2025GEN3C3W}. None of these methods models articulated hands, which in egocentric video largely hide the manipulated object, while depth stores only the nearest surface per ray. Multiple surfaces per ray are stored in layered depth images~\cite{Shade1998LayeredDI}, depth peeling~\cite{Everitt2001InteractiveOT}, and peeled human reconstruction~\cite{Jinka2020PeeledHumanRS}, whereas conditional generation conveys occlusion as instance ordering~\cite{Zhan2025LaRenderTO} or translucent 3D boxes~\cite{Agrawal2026SeeThrough3DOA}, revealing where hidden objects are but not their geometry; to our knowledge, multi-hit geometry has not conditioned video generation. \method renders the static scene as regular depth and the controlled meshes as peeled depth, carrying scene, camera, and action in one pixel-aligned signal whose raycast also masks our attention bias.
\section{Method}
\label{sec:method}
 
Our goal is to build an action-conditioned world model for dexterous manipulation: given a reference frame $\mathbf{I}_0$, a camera trajectory, and hand motion, object motion, or both, we predict the resulting egocentric video. Prior dexterous world models receive hand poses and camera motion as separate signals and must infer how the hands sit in the scene~\cite{Li2026EgocentricWM,Hao2026EgoSimEW,Xie2026GeneratedRH,Chen2026HandsOnWorldUE}. Object trajectory-conditioned video generators move objects along a given path, but they do not generate the hands that cause the motion or target egocentric camera motion~\cite{Fu20243DTrajMasterM3,Chu2025WanMoveMV}. \method instead rests on one design choice: every action, whatever its type, reaches the model as geometry rendered from the target camera. We formalize the task in \cref{sec:method:problem}, describe the geometric action compiler, which compiles scene, camera, hands, and objects into one control video, in \cref{sec:method:compiler}, and describe the generator and its training in \cref{sec:method:model,sec:method:training}.

\begin{figure}
    \centering
    \includegraphics[width=\linewidth]{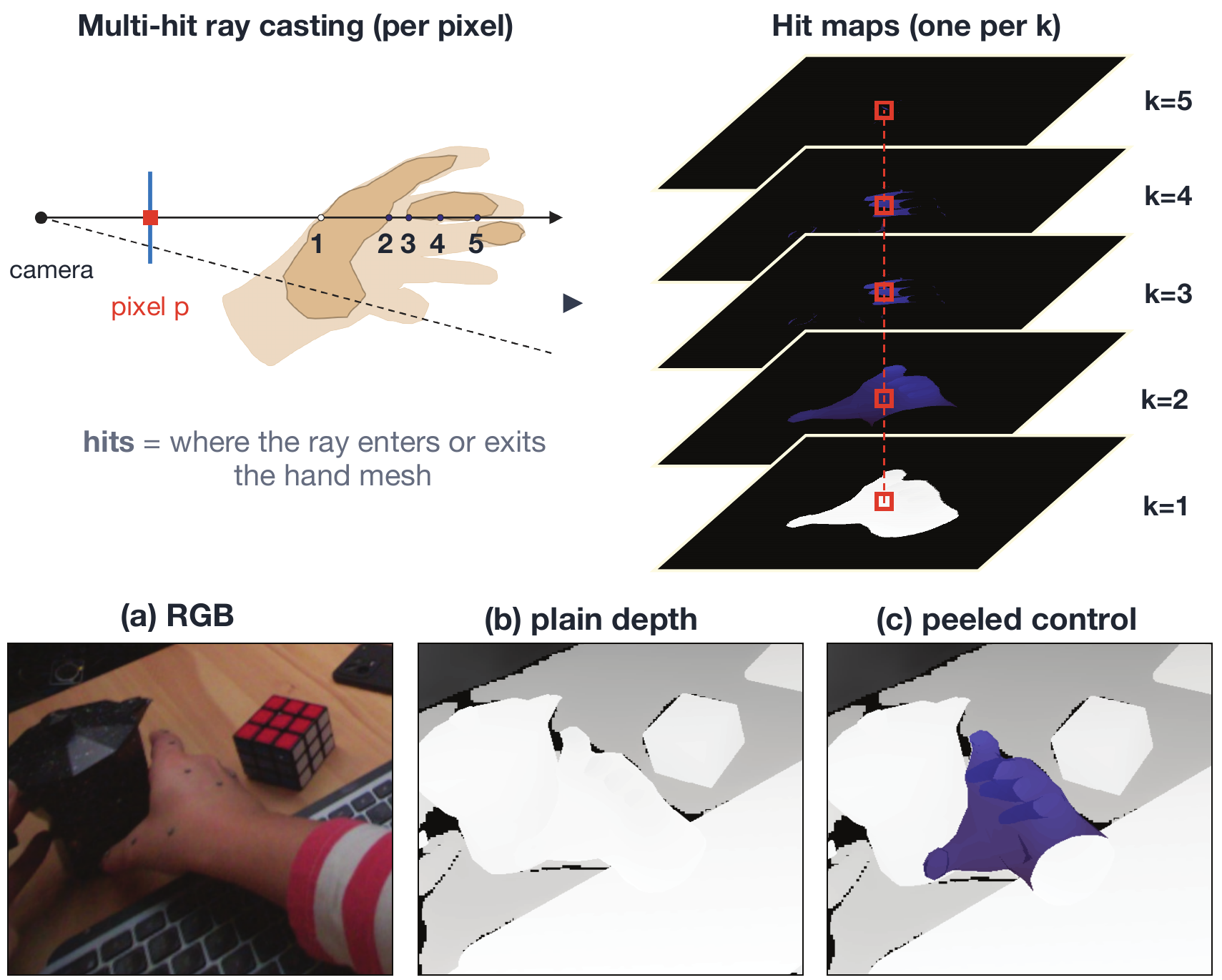}
    \vspace{-1.5em}
    \caption{\textbf{Peeled depth.} \emph{Top left:} for every pixel, we cast a ray from the camera against the controlled meshes (hands and/or object) and record each point where it enters or exits a mesh. \emph{Top right:} collecting the $k$-th hit of every pixel gives one map per $k$, colored by depth. $k{=}1$ is the visible hand surface, while deeper hits recover palm and fingers hidden behind other fingers or the grasped object. \emph{Bottom:} Cropped views of RGB scene (a), corresponding plain depth (b), and our peeled depth control (c), which composites all hits over the grayscale scene depth lifted from $\mathbf{I}_0$ (Eq.~\eqref{eq:control}), keeping the occluded fingers in the signal while leaving the scene in the plain depth format.}
    \label{fig:peel}
    \vspace{-1.0em}
\end{figure}

\subsection{Problem Formulation}
\label{sec:method:problem}
 
We are given a reference frame $\mathbf{I}_0$ with intrinsics $\mathbf{K}$, a camera trajectory $\{\mathbf{T}_t\}_{t=1}^{N}$, where $\mathbf{T}_t\in SE(3)$ maps the reference camera frame to the camera at frame $t$, and an action consisting of the posed MANO meshes of both hands, the 6-DoF trajectories of rigid objects with known meshes, or both. We denote the set of posed \emph{controlled} meshes at frame $t$ in the reference camera frame with $\mathcal{F}_t$. The control mode is defined entirely by which meshes $\mathcal{F}_t$ contains: in the hand mode it holds the hand meshes and the model must infer the object's response; in the object mode it holds the moving objects and the model must infer the hand motion and contacts; in the joint mode it holds both, the scene is omitted, and the model must also generate the scene and its appearance. In every mode, the model also completes the scene regions unseen in $\mathbf{I}_0$ that the moving camera reveals. We factor the model as:

\vspace{-1.5em}
\begin{equation}
\hat{\mathbf{V}}=G_\theta(\mathbf{I}_0,\mathbf{p},\mathbf{C}),\qquad \mathbf{C}=\mathcal{R}\big(\mathbf{I}_0,\mathbf{K},\{\mathbf{T}_t\},\{\mathcal{F}_t\}\big)
\label{eq:factor}
\end{equation}
\vspace{-1.25em}

where $\hat{\mathbf{V}}=\{\hat{\mathbf{I}}_t\}_{t=1}^{N}$ is the predicted video,  $\mathcal{R}$ is the fixed geometric action compiler that renders the inputs into a control video $\mathbf{C}=\{\mathbf{C}_t\}_{t=1}^{N}$ of three-channel frames, $\mathbf{p}$ is an optional text prompt, and $G_\theta$ is a video diffusion model. Since the control video is limited to the information available at deployment, the scene geometry comes from $\mathbf{I}_0$ alone, and additionally must preserve the occluded fingers and grasped objects that a depth map, which keeps only the nearest surface along each ray, discards.

\begin{figure*}[t]
    \centering
    \includegraphics[width=\textwidth]{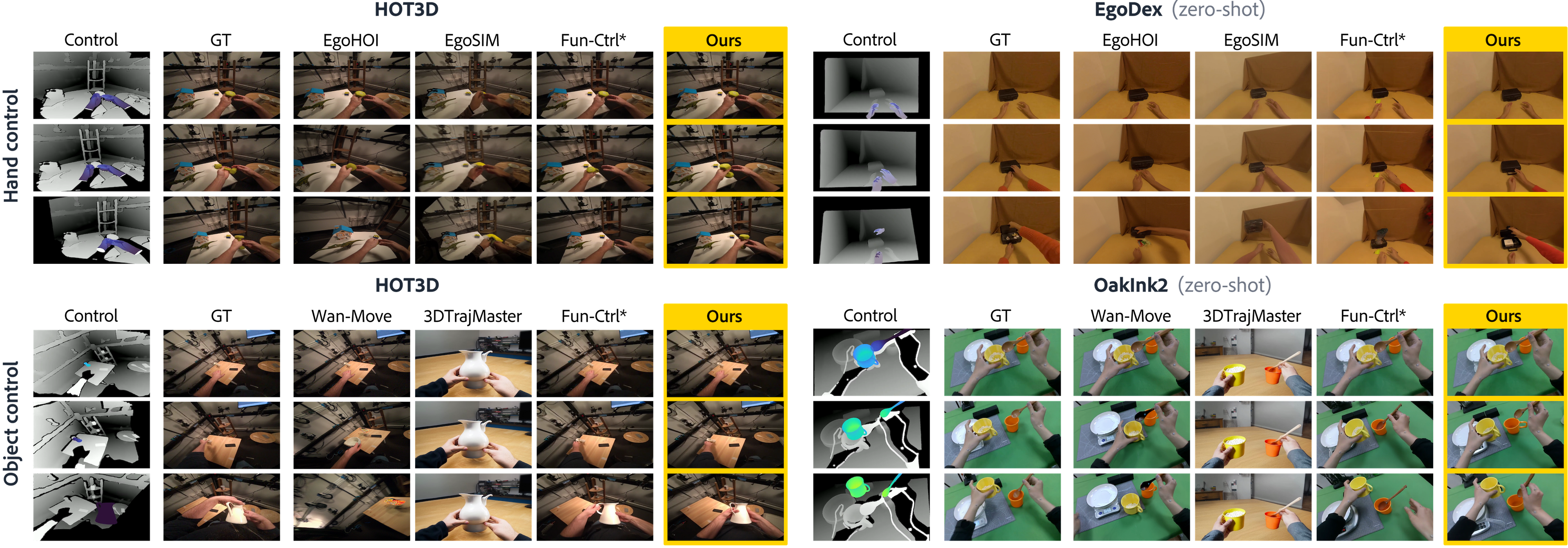}
    \vspace{-1.5em}
    \caption{\textbf{Qualitative comparison.} Each row shows the reference frame, our peeled control for our method, generated frames from \method and baselines, and ground truth. We evaluate hand-action (top) and object-action (bottom) conditioned results on the HOT3D test split, and zero-shot on EgoDex and OakInk2. \method better preserves both hand and object geometry and pose under self-occlusion and keeps camera motion aligned with the headset trajectory.}
    \label{fig:comparison}
\vspace{-1.0em}
\end{figure*}

\subsection{Geometric Action Compiler}
\label{sec:method:compiler}

Each control frame $\mathbf{C}_t$ consists of two layers rendered from camera $t$. The \emph{depth layer} is a grayscale depth map in the format the control branch was pretrained on and shows only the static scene lifted from $\mathbf{I}_0$. The \emph{peeled layer} is drawn on top in color and shows the controlled hands and objects, including their surfaces hidden from the camera.

\vspace{-.5em}
\paragraph{Depth layer.} We lift $\mathbf{I}_0$ to a point cloud $\mathcal{S}_0$ in the reference camera frame with Pi3X~\cite{Wang20253PV} using $\mathbf{K}$, remove points corresponding to hands segmented with SAM3~\cite{Carion2025SAM3S,Zeng2026SAM3LiteTextAA}, so that transient hand geometry does not get baked into the background, and discard low-confidence and depth-discontinuous points; objects visible in $\mathbf{I}_0$ remain at their initial poses. For each frame, we transform $\mathcal{S}_0$ by $\mathbf{T}_t$ and rasterize it into a metric depth map. Camera motion thus appears as the apparent motion of the scene rather than as a pose embedding, and regions unseen in $\mathbf{I}_0$ stay empty, marking where $G_\theta$ must generate content. We normalize the depth $Z_t(u)$ at pixel $u$ with one clip-level range $[z_{\min},z_{\max}]$, the 2nd and 98th percentiles of valid depths, as $L_t(u)=1-\operatorname{clip}\big((Z_t(u)-z_{\min})/(z_{\max}-z_{\min}),0,1\big)$, so that intensities are comparable across frames.

\vspace{-1.0em}
\paragraph{Peeled layer.} For every pixel $u$, we cast the camera ray against the controlled meshes $\mathcal{F}_t$ and record up to 12 ordered intersections, restarting $10^{-3}$\,m past each hit. The peeled set $\mathcal{P}_t(u)$ contains the visible surface of the controlled meshes and every surface occluded by the mesh itself, by another controlled mesh, or by the scene. Hands thus enter the control video only through the peeled layer, and a moving object is distinguished from its static copy in $\mathcal{S}_0$. We encode these hits in color, which depth control leaves unused since the three channels of a grayscale depth video are identical. Each hit $i$ with depth $d_i$, normalized with the same clip-level range, receives the color $\mathbf{c}_i=\operatorname{Turbo}(d_i)$, opacity $\alpha_i$, and weight $w_i$, with $\alpha_i=0.85$ and $w_i=\alpha_i[0.1+0.9(1-d_i)]$ for occluded hits, so that nearer hidden surfaces dominate, and $\alpha_i=w_i=1$ for visible hits. We then apply weighted blended order-independent transparency~\cite{McGuire2013WeightedBO} to composite the hits over the depth layer without sorting:

\vspace{-1.5em}
\begin{equation}
\mathbf{C}_t(u)=\big(1-a_t(u)\big)\,L_t(u)\,\mathbf{1}
+a_t(u)\,\frac{\sum_{i\in\mathcal{P}_t(u)}w_i\,\mathbf{c}_i}{\sum_{i\in\mathcal{P}_t(u)}w_i},
\label{eq:control}
\end{equation}
\vspace{-1em}

where $\mathbf{1}$ replicates $L_t$ across the three channels and $a_t(u)=1-\prod_{i\in\mathcal{P}_t(u)}(1-\alpha_i)$ is the accumulated coverage, which is zero when $\mathcal{P}_t(u)$ is empty. Pixels without peeled hits therefore remain plain depth, in distribution for the pretrained branch, and the control mode is conveyed solely by which meshes appear peeled, without a task identifier or action-specific encoder.

\begin{table*}[t]
\caption{\textbf{Hand-controlled generation on HOT3D and EgoDex (zero-shot).} Video quality (PSNR in dB), hand kinematics from WiLoR, and camera motion from MapAnything after Sim(3) alignment. Fun Ctrl* denotes our base model fine-tuned on HOT3D with depth-only control videos. Best results are in \textbf{bold}.}
\vspace{-1.5em}
\label{tab:metrics}
\begin{center}
\resizebox{0.85\linewidth}{!}{%
\begin{tabular}{c l ccc cc ccc}
\toprule
& \textbf{Method}
  & \multicolumn{3}{c}{\textbf{Video quality}}
  & \multicolumn{2}{c}{\textbf{Hand kinematics}}
  & \multicolumn{3}{c}{\textbf{Camera motion}} \\
\cmidrule(lr){3-5} \cmidrule(lr){6-7} \cmidrule(lr){8-10}
&
  & \textbf{PSNR $\uparrow$}
  & \textbf{SSIM $\uparrow$}
  & \textbf{LPIPS $\downarrow$}
  & \textbf{MPJPE $\downarrow$}
  & \textbf{PCK@20 $\uparrow$}
  & \textbf{ATE $\downarrow$}
  & \textbf{RRE $\downarrow$}
  & \textbf{RPE$_{\mathrm{rot}}$ $\downarrow$} \\
\midrule
\multirow{4}{*}{\rotatebox[origin=c]{90}{\textbf{HOT3D}}}
& Wan2.2 I2V~\cite{Wang2025WanOA}
  & 13.56 &  0.458 & 0.546
  & 97.73 & 15.5
  & 0.158 & 114.0 & 1.57 \\
& Fun-Ctrl*~\cite{aigc_apps_VideoX_Fun_2026}
  & \cII 16.27 &  0.514 & \cII 0.320
  &  37.92 &  63.3
  & \cII 0.050 & \cII 55.2 & \cII 0.74 \\
& EgoHOI~\cite{Li2026EgocentricWM}
  & 14.23 & 0.457 &  0.541
  & \cII 16.24 & \cII 81.8
  & 0.179 & 118.5 &  0.99 \\
& EgoSIM~\cite{Hao2026EgoSimEW}
  &  15.33 & \cII 0.520 &  0.467
  & 48.74 & 48.3
  &  0.058 &  81.6 &  0.81 \\
& \ours Ours
  & \cI \textbf{19.01} & \cI \textbf{0.686} & \cI \textbf{0.264}
  & \cI \textbf{14.26} & \cI \textbf{83.8}
  & \cI \textbf{0.045} & \cI \textbf{26.6} & \cI \textbf{0.67} \\
\midrule
\multirow{5}{*}{\rotatebox[origin=c]{90}{\textbf{EgoDex}}}
& Wan2.2 I2V~\cite{Wang2025WanOA}
  & 19.17 & \cII 0.798 & \cII 0.359
  & 60.91 & 37.7
  & 0.034 & 100.6 & 0.93 \\
& Fun-Ctrl*~\cite{aigc_apps_VideoX_Fun_2026}
  & 18.10 & 0.781 & 0.385
  & 41.03 & 55.6
  &  0.016 & 107.0 &  0.62 \\
& EgoHOI~\cite{Li2026EgocentricWM}
  &  19.30 &  0.795 & 0.363
  &  27.43 &  63.5
  & \cII 0.014 &  100.4 & 0.79 \\
& EgoSIM~\cite{Hao2026EgoSimEW}
  & \cII 20.27 & 0.794 &  0.360
  & \cII 21.77 & \cII 71.2
  & 0.021 & \cII 88.0 & \cII 0.47 \\
& \ours Ours
  & \cI \textbf{21.83} & \cI \textbf{0.836} & \cI \textbf{0.302}
  & \cI \textbf{20.38} & \cI \textbf{77.3}
  & \cI \textbf{0.013} & \cI \textbf{63.3} & \cI \textbf{0.38} \\
\bottomrule
\end{tabular}
}
\end{center}
\vspace{-1.5em}
\end{table*}

\subsection{Dexterous Rollout Generator}
\label{sec:method:model}

We initialize $G_\theta$ from Wan2.2-Fun-A14B-Control~\cite{Wang2025WanOA}, a video DiT conditioned on a reference image, a text prompt, and a three-channel control video, and only replace its depth input with our control video. This control video is deliberately incomplete: it omits regions unseen in $\mathbf{I}_0$, all appearance, and the uncontrolled entity, \ie, the object under hand control and the hands under object control. Because each partial control video is paired with the full real video during training, the flow-matching objective alone teaches $G_\theta$ to complete these parts consistently with the action, acting as a world model rather than as a renderer of its control.

\vspace{-1.0em}
\paragraph{Action-biased attention.}
\label{sec:method:attn}
Everything $G_\theta$ completes depends on the action: an object already grasped in the first frame must adhere to the motion of the hand holding it, and generated hands must complement the target object motion given. Yet the controlled meshes usually cover a small fraction of each frame, so their tokens compete with many more scene tokens, and we therefore bias attention toward them. The raycast yields a mask of the \emph{visible} controlled surfaces, excluding occluded hits since they lie on pixels owned by an occluder, which we pool to the latent token grid to label each video token as foreground ($f=1$) or background ($f=0$); everything $G_\theta$ must complete is thus background, and reference-image tokens are treated as background keys. For a query $q$ and key $k$, with $b_q=1-f_q$, we add a bias to the pre-softmax logit $A^{\ell,h}_{qk}$ of every self-attention layer $\ell$ and head $h$:

\vspace{-1em}
\begin{equation}
\widetilde{A}^{\ell,h}_{qk}=A^{\ell,h}_{qk}+g_{\ell,h}\,\gamma(\tau)\,b_q\,f_k
\label{eq:bias}
\end{equation}
\vspace{-1.0em}

where the learnable scalars $g_{\ell,h}$ are initialized to zero, so that training starts from the base model, and $\gamma(\tau)=\operatorname{clip}(\tau/1000,0,1)$ applies the bias most strongly at high noise levels $\tau$, when coarse structure is determined. The factor $b_q$ makes the bias asymmetric: background queries attend more to the action, while attention among foreground tokens, already constrained by the peel, is unchanged. We implement \cref{eq:bias} as a FlexAttention score modifier; the bias adds 1.8 dB PSNR over peeling alone (\cref{tab:ablation}).

\subsection{Training}
\label{sec:method:training}

We train a single model on hand- and object-controlled clips with the flow-matching objective of the base model, freezing the VAE and text encoder and fine-tuning the DiT with LoRA~\cite{Hu2021LoRALA} of rank 32 on the query, key, value, and output projections of all self- and cross-attention layers and on the feed-forward layers, together with the gates $g_{\ell,h}$. We optimize with AdamW at an initial learning rate of $10^{-4}$ with a cosine schedule and a batch size of 1 for 40k steps, on clips of 49 frames at 16~fps and $832\times480$ resolution captioned with Qwen3.5-9B. We drop the text and reference image independently with probability $0.3$, which allows generation from geometric control alone, and additionally include scene-free joint controls for the sim-to-real application in \cref{sec:exp:setting3}. The same model, without mode-specific parameters or objectives, is used in all experiments.

\section{Experiments}
\label{sec:exp}

We evaluate \method as a dexterous world model under hand control and object control (\cref{sec:exp:setting1,sec:exp:setting2}), and then study two downstream uses enabled by the same control video interface (\cref{sec:exp:apps}). All results use a single model trained only on HOT3D~\cite{Banerjee2024HOT3DHA}; we provide additional implementation details, qualitative results and ablations in the supplementary materials.

\begin{table*}[t]
\caption{\textbf{Object-controlled generation on HOT3D and OakInk2 (zero-shot).} We evaluate object motion adherence using FoundationPose, realism of the inferred hands via WiLoR statistics, and distributional video quality (FVD, FID). *Fine-tuned on HOT3D with depth-only control videos.}
\vspace{-1.5em}
\label{tab:object_metrics}
\begin{center}
\resizebox{0.85\linewidth}{!}{%
\begin{tabular}{c lccc c cc }
\toprule
& \textbf{Method}
& \multicolumn{3}{c}{\textbf{Object motion}}
& \multicolumn{1}{c}{\textbf{Hand realism}}
& \multicolumn{2}{c}{\textbf{Video quality}} \\
\cmidrule(lr){3-5} \cmidrule(lr){6-6} \cmidrule(lr){7-8}
&
& \textbf{ADD-S (cm) $\downarrow$}
& \textbf{RotErr (deg) $\downarrow$}
& \textbf{DetRate (\%) $\uparrow$}
& \textbf{WiLoR stat.\ $\downarrow$}
& \textbf{FVD $\downarrow$}
& \textbf{FID $\downarrow$} \\
\midrule
\multirow{4}{*}{\rotatebox[origin=c]{90}{\textbf{HOT3D}}}
& Wan-Move~\cite{Chu2025WanMoveMV}
&  25.00 &  58.30 & 84.79 &  7.12 &  252.85 &  29.47 \\
& 3DTrajMaster~\cite{Fu20243DTrajMasterM3}
& 33.75 & 75.12 & \cII 90.55 & 11.35 & 2056.07 & 169.01 \\
& Fun-Ctrl*~\cite{aigc_apps_VideoX_Fun_2026}
& \cII 5.83 & \cII 26.90 &  90.05 & \cII 6.28 & \cII 140.55 & \cII 16.24 \\
& \ours Ours
& \cI \textbf{2.66} & \cI  \textbf{16.77} & \cI \textbf{95.12} & \cI \textbf{4.89} & \cI \textbf{97.36} & \cI \textbf{12.42} \\
\midrule
\multirow{4}{*}{\rotatebox[origin=c]{90}{\textbf{OakInk2}}}
& Wan-Move~\cite{Chu2025WanMoveMV}
&  19.79 &  53.55 & 88.00 & \cII 5.74 &  460.21 &  38.78 \\
& 3DTrajMaster~\cite{Fu20243DTrajMasterM3}
& 31.78 & 78.20 & \cII 91.05 & 12.37 & 2782.80 & 235.15 \\
& Fun-Ctrl*~\cite{aigc_apps_VideoX_Fun_2026}
& \cII 5.46 & \cII 22.81 & \cII 92.38 &  5.86 & \cII 336.20 & \cII 37.28 \\
& \ours Ours
& \cI \textbf{2.10} & \cI \textbf{14.94} & \cI \textbf{96.12} & \cI \textbf{4.82} & \cI \textbf{228.21} & \cI \textbf{13.57} \\
\bottomrule
\end{tabular}
}
\end{center}
\vspace{-2em}
\end{table*}

\subsection{Experimental Setup}
\label{sec:exp:setup}

\paragraph{Datasets.} HOT3D provides egocentric Aria recordings with headset poses, 3D MANO and UmeTrack hand trajectories, and posed CAD object models in a shared metric frame. Since HOT3D's official test set does not have hand poses, we create a custom train-test split by holding out 12 entire recording sequences for evaluation and train on the remainder. We convert all HOT3D clips from fisheye to pinhole, sample them as 92-frame windows at 30~fps with 16-frame overlap, and resample them to 49 frames at 16~fps via frame dropping (\cref{sec:method:training}), resulting in 6380 training and 632 test samples. To test generalization, we additionally evaluate hand control on EgoDex~\cite{Hoque2025EgoDexLD} and object control on OakInk2~\cite{Zhan2024OakInk2A} zero-shot using 500 samples each from their respective test sets. Like HOT3D, OakInk2 is a bimanual hand-object manipulation dataset with egocentric and exocentric videos, 3D MANO hand trajectories, and posed CAD object models. EgoDex provides only ARKit keypoints, so we fit MANO hands with HaWoR~\cite{Zhang2025HaWoRWH} and use the same fitted trajectories as input to all methods.

\vspace{-1em}
\paragraph{Baselines.} For hand control, we compare against state-of-the-art dexterous world models with open source weights: EgoHOI~\cite{Li2026EgocentricWM}, which is trained on HOT3D and may have seen our held-out sequences, and EgoSIM~\cite{Hao2026EgoSimEW}, and the image-to-video model Wan2.2 I2V~\cite{Wang2025WanOA}. For object control, we compare against the object-trajectory video generators Wan-Move~\cite{Chu2025WanMoveMV} and 3DTrajMaster~\cite{Fu20243DTrajMasterM3}. In both settings, we also report Wan2.2 Fun-Control (denoted as Fun-Ctrl*)~\cite{aigc_apps_VideoX_Fun_2026}, the same base model as ours fine-tuned on the same HOT3D split with control videos that contain only the scene and overlaid action depth layers, without peeling or the action-biased attention.

\vspace{-1em}
\paragraph{Evaluation protocol and metrics.} For each test clip, we use the first frame as $\mathbf{I}_0$ and a fixed 3\,s temporal window with identical start and end times for every method. \method generates $832\times480$ video at 16~fps, whereas each baseline receives the same reference frame and the action information its native interface accepts, and runs with its released checkpoint, inference script, preprocessing steps, and native frame rate and resolution; we refer to the supplementary material for additional details. Our control videos never use target RGB, future scene geometry, or depth beyond what is lifted from the first frame $\mathbf{I}_0$ (\cref{sec:method:compiler}). In object control, we treat an object as controlled when its world-frame pose changes by more than 2~cm or $5^\circ$ within the clip.

We report PSNR, SSIM, and LPIPS against ground-truth RGB. We measure hand fidelity by running WiLoR~\cite{Potamias2024WiLoRE3} on generated and real videos and reporting MPJPE (mm) and PCK@20mm (\%) after Procrustes alignment. We measure object fidelity via FoundationPose~\cite{Wen2023FoundationPoseU6} as ADD-S (cm), geodesic rotation error (deg), and detection rate (\%). Camera motion is estimated with MapAnything~\cite{Keetha2025MapAnythingUF} and evaluated after Sim(3) alignment using ATE (m), RRE (deg), and rotational RPE (deg). For object control, we additionally report FVD and FID, and measure the realism of the inferred hands as the $\ell_2$ distance between WiLoR kinematic statistics of generated and ground-truth videos. As these geometric metrics rely on fixed estimators, we compare estimates on generated and ground-truth videos and treat them as relative indicators of fidelity; against ground-truth annotations on HOT3D, the estimators' noise floors are 5.8~mm MPJPE and 98.9\% PCK@20 for WiLoR, and 2.10~cm ADD-S and 12.3$^\circ$ rotation error for FoundationPose.

\subsection{Hand-Action Conditioned Rollouts}
\label{sec:exp:setting1}
We first evaluate whether \method follows given hand and camera trajectories while preserving the appearance of reference frame $\mathbf{I}_0$ and present results in \cref{tab:metrics}. This is the counterfactual query a visual simulator or planner must answer: what would this scene look like if the hands executed \emph{this} trajectory? On HOT3D and, zero-shot, on EgoDex, \method matches or outperforms all baselines in video quality, hand kinematics, and camera motion. On HOT3D, it improves PSNR from 15.33 to 19.01~dB over the strongest dexterous world model, EgoSIM, reduces LPIPS from 0.467 to 0.264, and lowers camera RRE from $81.6^\circ$ to $26.6^\circ$, while achieving the lowest MPJPE and highest PCK@20, surpassing the most accurate hand baseline, EgoHOI. Notably, our Fun-Ctrl* baseline, which already renders scene, camera, and hands into a single depth video, is competitive with the dedicated dexterous world models, particularly in camera motion. We attribute this to the unified control video, which places scene, camera, and hands in a shared 3D frame instead of feeding them through separate encoders; the peeled layer and action-biased attention account for the remaining gap (\cref{sec:exp:ablation}). Qualitative comparisons are shown in \cref{fig:comparison}.

\subsection{Object-Action Conditioned Rollouts}
\label{sec:exp:setting2}

We next ask whether the same model remains accurate when only the object trajectory is given and the hands must be inferred. We compare against specialized SOTA object trajectory-conditioned video generators on HOT3D and, zero-shot, on OakInk2, both of which provide CAD models and ground-truth 6-DoF object trajectories. As shown in \cref{tab:object_metrics}, \method outperforms all baselines in object motion fidelity, hand realism, and video quality. On HOT3D, it reduces ADD-S from 25.00~cm for Wan-Move and 5.83~cm for Fun-Ctrl* to 2.66~cm, while improving the detection rate and the realism of the inferred hands. The same trend holds on OakInk2, where \method again achieves the lowest object errors and FVD despite never training on it. We attribute the largest gains over trajectory-specialized baselines to peeled depth retaining object surfaces hidden by the hand or scene, which plain depth or 2D tracks cannot represent; we include stratified results by rotation magnitude and occlusion level in the supplementary material.

\subsection{Applications}
\label{sec:exp:apps}
\vspace{-1em}
\begin{table}[h]
\caption{\textbf{Sim-to-real HOI generation.} Scene-free Dex2HOI control videos without a reference image; we report hand (MPJPE, PCK@20 (\%)) and object metric (ADD-S, rotation error) for generated videos, and VBench smoothness and consistency.}
\vspace{-1.5em}
\label{tab:sim2real_hoi}
\begin{center}
\resizebox{\linewidth}{!}{%
\small
\setlength{\tabcolsep}{2.3pt}
\begin{tabular}{@{}l cc cc cc@{}}
\toprule
& \multicolumn{2}{c}{\textbf{Hand}} & \multicolumn{2}{c}{\textbf{Object}} & \multicolumn{2}{c}{\textbf{VBench}} \\
\cmidrule(lr){2-3} \cmidrule(lr){4-5} \cmidrule(l){6-7}
\textbf{Method} & \textbf{MPJPE$\downarrow$} & \textbf{PCK$\uparrow$} & \textbf{ADD-S$\downarrow$} & \textbf{Rot.$\downarrow$} & \textbf{Smooth.$\uparrow$} & \textbf{Cons.$\uparrow$} \\
\midrule
Fun-Ctrl*~\cite{aigc_apps_VideoX_Fun_2026} & 44.00 & 56.1 & 14.90 & 30.74 & 97.52 & 21.01 \\
\ours Ours & \cI \textbf{27.88} & \cI \textbf{64.1} & \cI \textbf{8.82} & \cI \textbf{24.54} & \cI \textbf{97.82} & \cI \textbf{23.45} \\
\bottomrule
\end{tabular}
}
\end{center}
\vspace{-2.5em}
\end{table}

\paragraph{Sim-to-real HOI generation.}
\label{sec:exp:setting3}
We test whether \method can turn simulated, scene-free interactions into realistic video with imagined scenes using the joint mode. We generate bimanual hand-object interactions with Dex2HOI~\cite{Pratikaki2026Dex2HOIDB} from text prompts derived from GRAB~\cite{GRAB:2020} metadata and enriched with Qwen3.5-9B, render them as control videos containing only the peeled hands and objects, and generate video from the control video and text alone, without a reference image. Since Dex2HOI outputs no camera, we place a static camera between the GRAB eye keypoints with an approximate gaze direction to mimic an egocentric view. As shown in \cref{tab:sim2real_hoi}, \method improves the hand and object kinematics recovered from the generated videos, as well as VBench~\cite{huang2023vbench, zheng2025vbench2} temporal smoothness and visual consistency, relative to Fun-Ctrl* given the same control videos.

\vspace{-1em}
\paragraph{Dexterous retargeting.}
We close the loop from video generation to robot execution. Do as I Do~\cite{Paliwal2026DoAI} reconstructs hand and object motion from \emph{real} egocentric videos and retargets it to a 22-DoF Sharpa Wave hand in MuJoCo and on hardware; we use it unchanged and only replace its input video with our \method's rollouts. On OakInk2, where \method runs zero-shot, we follow the protocol of~\cite{Paliwal2026DoAI}: each row of \cref{tab:retargeting} takes one modality from ground truth and reconstructs the other from either the generated or the corresponding real video. Under hand control, this tests whether the generated object motion is retargetable; under object control, the planning case of interest, Do as I Do must recover from video the hand motion that executes a given 6-DoF object trajectory. Success rates and pose errors from generated videos remain within 3–6 percentage points of those from real videos, indicating that \method rollouts can replace real video as input to a retargeting pipeline designed for real observations.

\begin{figure}
    \centering
    \includegraphics[width=0.95\linewidth]{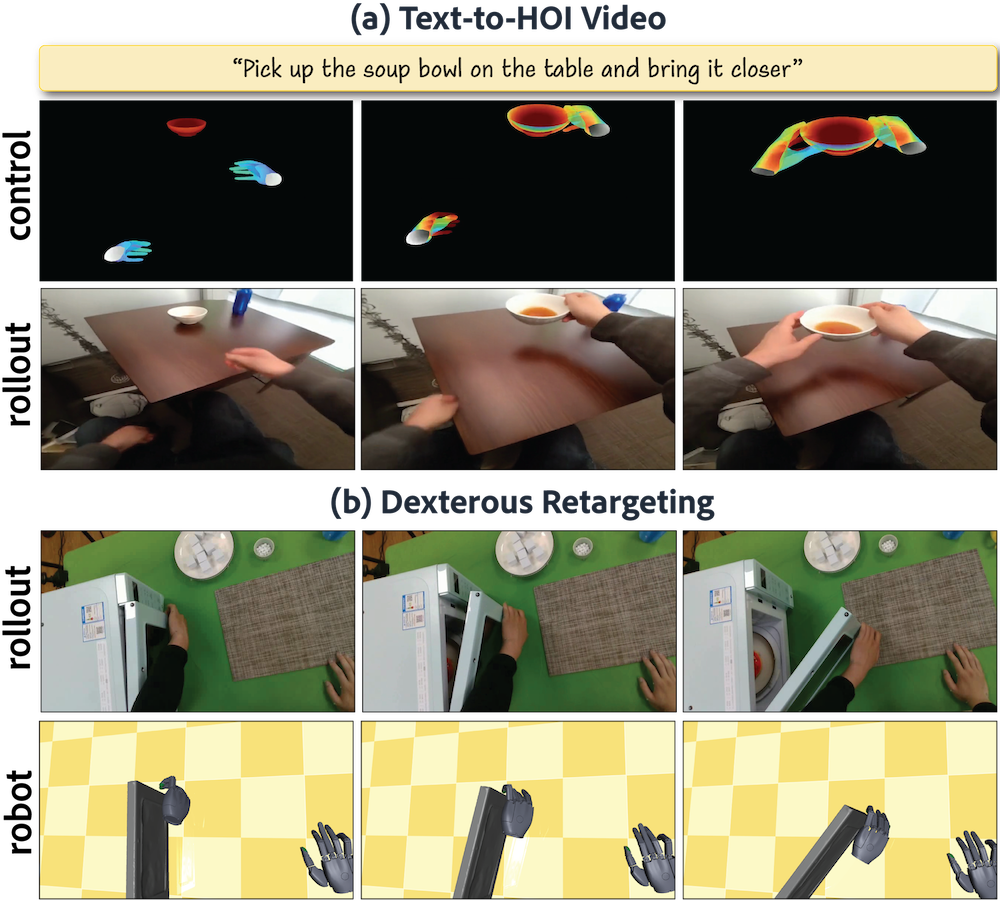}
    \vspace{-.5em}
    \caption{\textbf{Applications.} \emph{Top:} we use Dex2HOI to generate parametric HOI from text, convert it to a peeled control video and generate a realistic video rollout with \method. \emph{Bottom:} We generate an object-trajectory conditioned rollout with \method and then use Do as I Do to retarget it to a Sharpa hand in MuJoCo. }
    \label{fig:applications}
    \vspace{-1.5em}
\end{figure}

\begin{table}[t]
\caption{\textbf{Dexterous retargeting from generated \emph{vs.}\ real videos.} Do as I Do~\cite{Paliwal2026DoAI} is applied without retraining; each modality is taken from ground truth (GT) or reconstructed from video (Rec.). We report success rate (\%) and mean position (m) and rotation (rad) errors; a trial succeeds if they are below $0.1$\,m and $0.5$\,rad.}
\vspace{-1.5em}
\label{tab:retargeting}
\begin{center}
\resizebox{\columnwidth}{!}{%
\small
\setlength{\tabcolsep}{3.4pt}
\begin{tabular}{@{}ll cc ccc@{}}
\toprule
& & \multicolumn{2}{c}{\textbf{Reference}} & \multicolumn{3}{c}{\textbf{Retargeting}} \\
\cmidrule(lr){3-4} \cmidrule(l){5-7}
\textbf{Setting} & \textbf{Video} & \textbf{Hands} & \textbf{Object} & \textbf{Succ.$\uparrow$} & \textbf{Pos.$\downarrow$} & \textbf{Rot.$\downarrow$} \\
\midrule
-- & -- & GT & GT & 88 & 0.02 & 0.12 \\
\midrule
\multirow{2}{*}{Hand ctrl.} & Real & GT & Rec. & 75 & 0.05 & 0.21 \\
& Ours & GT & Rec. & 72 & 0.06 & 0.29 \\
\midrule
\multirow{2}{*}{Obj.\ ctrl.} & Real & Rec. & GT & 80 & 0.04 & 0.18 \\
& Ours & Rec. & GT & 77 & 0.05 & 0.22 \\
\midrule
\multirow{3}{*}{Full rec.} & Real & Rec. & Rec. & 74 & 0.05 & 0.25 \\
& Ours (hand) & Rec. & Rec. & 72 & 0.05 & 0.27 \\
& Ours (object) & Rec. & Rec. & 68 & 0.07 & 0.31 \\
\bottomrule
\end{tabular}
}
\end{center}
\vspace{-2em}
\end{table}

\subsection{Ablation Studies}
\label{sec:exp:ablation}
We ablate the two components that \method adds to depth-only control on the HOT3D test split under the hand- and object-control protocols of \cref{sec:exp:setting1,sec:exp:setting2}, fine-tuning all variants from Wan2.2-Fun-A14B-Control with the same schedule (\cref{sec:method:training}). In \cref{tab:ablation}, we report a cumulative ladder: depth-only control (Fun-Ctrl* in \cref{tab:metrics,tab:object_metrics}), the added peeled layer, and the full model with the attention bias of \cref{eq:bias}. We provide ablations of scene lifting, attention-bias variants, and the peel encoding in the supplementary materials.

As shown in \cref{tab:ablation}, the peeled layer is the main driver of fidelity in both modes: it roughly halves MPJPE and RRE under hand control and accounts for about three quarters of the ADD-S reduction under object control, consistent with its role in retaining hand and object surfaces hidden by occlusion. The attention bias further improves video quality, hand kinematics, and object pose, reducing ADD-S from 3.43 to 2.66~cm, while leaving camera motion essentially unchanged, indicating that it mainly refines the appearance and completion of the interaction rather than egomotion.

\begin{table}[h]
\caption{\textbf{Component ablation on HOT3D.} Each component is added cumulatively; Depth only* is Fun-Ctrl* in \cref{tab:metrics,tab:object_metrics}.}
\vspace{-1.5em}
\label{tab:ablation}
\begin{center}
\resizebox{\columnwidth}{!}{%
\small
\setlength{\tabcolsep}{3.5pt}
\begin{tabular}{@{}l ccc cc@{}}
\toprule
& \multicolumn{3}{c}{\textbf{Hand control}} & \multicolumn{2}{c}{\textbf{Object control}} \\
\cmidrule(lr){2-4} \cmidrule(l){5-6}
\textbf{Control} & \textbf{PSNR$\uparrow$} & \textbf{MPJPE$\downarrow$} & \textbf{RRE$\downarrow$} & \textbf{ADD-S$\downarrow$} & \textbf{Rot.$\downarrow$} \\
\midrule
Depth only* & 16.27 & 37.92 & 55.2 & 5.83 & 26.90 \\
+ Peeled layer & \cII 17.19 & \cII 17.99 & \cI \textbf{25.4} & \cII 3.43 & \cII 20.00 \\
\ours + Fg.\ bias (full) & \cI \textbf{19.01} & \cI \textbf{14.26} & \cII 26.6 & \cI \textbf{2.66} & \cI \textbf{16.77} \\
\bottomrule
\end{tabular}
}
\end{center}
\vspace{-1.5em}
\end{table}
\section{Conclusion}
\label{sec:conclusion}

We introduced \method, a dexterous world model that can be driven by bimanual hand trajectories, 6-DoF object trajectories, or both within a single model. By compiling scene, camera, and action into one control video, with the scene lifted from the reference frame and the controlled meshes rendered as peeled depth, \method conditions a pretrained depth-controlled video model to follow every control mode without action-specific encoders, outperforming specialized baselines in both modes and transferring zero-shot to unseen datasets. Limitations include the reliance on first-frame scene reconstruction and on known hand and object meshes; extending the interface to articulated and deformable objects, and decoding inferred trajectories, are natural next steps.

\paragraph{LLM Usage Statement.} We used Qwen3.5-9B to caption videos and enrich existing dataset captions where applicable. We additionally used Google Deep Review to identify issues and suggest supporting experiments, and Claude Opus 5.5 to lightly edit \cref{sec:intro,sec:method,sec:conclusion} for clarity.

{
\small
\bibliographystyle{ieeenat_fullname}
\bibliography{main}
}

\appendix
\providecommand{\todo}[1]{\textcolor{red}{[TODO: #1]}}

\section{Implementation Details}
\label{sec:supp:impl}

\subsection{Geometric action compiler}
\label{sec:supp:compiler}

\paragraph{Scene layer.} We build the point cloud of the scene $\mathcal{S}_0$ with the same procedure for training and for every set of evaluations without using the geometry of the ground-truth scene. More specifically, we use Pi3X \cite{Wang20253PV} to lift the first frame $\mathbf{I}_0$ with intrinsics $\mathbf{K}$, and we discard points with confidence below $0.1$ or on depth discontinuities, detected with a relative threshold of $0.03$. We remove the hands in the scene with SAM3-LiteText \cite{Zeng2026SAM3LiteTextAA} so that their geometry does not become part of the static scene, prompting with ``hands'' at confidence $0.35$ in HOT3D \cite{Banerjee2024HOT3DHA}, OakInk2 \cite{Zhan2024OakInk2A} and EgoDex \cite{Hoque2025EgoDexLD}. Objects visible in $\mathbf{I}_0$ remain in $\mathcal{S}_0$ at their initial pose, so their subsequent motion appears only in the peeled layer.

\paragraph{Scene scale.} We express the scene layer and the action meshes in a common metric frame. On HOT3D and OakInk2, we anchor $\mathcal{S}_0$ to the metric hand and controlled-object meshes, projecting the first-frame vertices into $\mathbf{I}_0$ and rescaling $\mathcal{S}_0$ by the median ratio of vertex depth to Pi3X depth over vertices within $(0.05, 3)$\,m that land on valid points before hand removal, keeping $s=1$ for clips with fewer than $50$ such vertices. On EgoDex, which provides ARKit keypoints rather than MANO, we fit MANO with HaWoR \cite{Zhang2025HaWoRWH} and condition all methods on the same fits (\cref{tab:supp:hawor}). HaWoR recovers the scale of its SLAM camera trajectory by aligning it with Metric3D \cite{Yin2023Metric3DTZ} estimated monocular metric depth; we perform this alignment with Pi3X instead, which places the reconstructed hands in the same units as $\mathcal{S}_0$.

\begin{table}[h]
\centering
\small
\caption{\textbf{HaWoR fits on EgoDex.} Mean residual between ARKit keypoints and their corresponding MANO joints after root alignment. We discard clips with a residual above 7.5 mm and evaluate all methods on the remaining clips.}
\label{tab:supp:hawor}
\begin{tabular}{@{}lc@{}}
\toprule
\textbf{Statistic} & \textbf{Value} \\
\midrule
Mean keypoint residual (mm) & 4.84 \\
Clips discarded (\%) & 21.1 \\
\bottomrule
\end{tabular}
\end{table}

\subsection{Dexterous rollout generator}
\label{sec:supp:generator}

\paragraph{Action-biased attention.} We rasterize the action mask $\mathbf{M}\in\{0,1\}^{T\times H\times W}$ from the same raycast that builds the peel, setting $M_{t,h,w}=1$ where the first hit belongs to a controlled surface (hands, objects, or both, depending on the mode). The mask enters only the bias in Eq. 3 of the main paper; the controlled motion reaches $G_\theta$ through the peeled layer. We max-pool $\mathbf{M}$ over the support of each DiT token, which spans $16\times16$ pixels after the $8\times$ VAE downsampling and the $(1,2,2)$ patch embedding, and the frames the causal VAE groups into one latent frame, \ie frame $0$ alone and frames $\{4f'{-}3,\ldots,4f'\}$ for $f'\geq1$. A token is thus foreground ($f_i=1$) when any pixel in its support is foreground, and we set $b_q=1-f_q$. Reference-image tokens receive $f=0$ and act as background keys, so the bias acts only on background-query--foreground-key pairs and leaves all other pairs unchanged.

\paragraph{Sim-to-real HOI generation.} We use the official repository with the default settings and GRAB \cite{GRAB:2020} checkpoint of Dex2HOI \cite{Pratikaki2026Dex2HOIDB}, which is a text-to-parametric HOI sequence method that is trained on single-object HOI data consisting of triplets of a text description of the action, SMPL-X \cite{Pavlakos2019ExpressiveBC} full body trajectories, and a 6-DoF trajectory for the CAD object manipulated. For our sceneless generation experiments, we use the test split of the GRAB dataset consisting of 145 samples, enrich the existing text descriptions with Qwen3.5-9B \cite{qwen3.5}. For each enriched text description and target CAD object, we generate a parametric full-body HOI sequence consisting of 60 frames. To create the peeled control videos for our method, we place the virtual egocentric camera at the midpoint of the SMPL-X eye joints, with a $40^\circ$ downward pitch and gaze toward the wrists and manipulated object, retain only the generated CAD object and the bimanual hand meshes (MANO topology) extracted from the SMPL-X body, and otherwise follow our geometric action compiler protocol outlined in Sec.~3.2 of our main paper without a scene depth layer, rendering both the object and bimanual hands motion as a peeled layer on black background. Finally, we use \method without a reference image to perform sceneless HOI to video generation, which is essentially a sim-to-real task. We provide additional video samples for this mode in the attached video folder.

\paragraph{Data preprocessing.} We train only on our HOT3D training split and evaluate on the held-out test split sequences listed in \cref{tab:test-set}. We assign 50\% of the training clips to hand control and the other 50\% to object control, so that the model never sees the same target video under two control modes, and we additionally render a random 10\% of all training clips without the scene layer for joint hand-object control. For hand control, we discard clips in which no hand is visible in more than 80\% of the frames. For object control, we treat every object that translates by more than 2\,cm or rotates by more than $5^\circ$ from its initial pose as controlled and render all of them in the peeled layer, so a control video can contain several controlled objects; we discard clips without any such object. We apply the same filters to the evaluation clips, the hand filter on EgoDex and the object filter on OakInk2.

\begin{table}[t]
\centering
\caption{\textbf{HOT3D test split.} The 12 long Aria sequences we hold out for evaluation; we train on all remaining HOT3D Aria sequences. Each sequence consists of multiple demonstrations with multiple objects.}
\label{tab:test-set}
\small
\setlength{\tabcolsep}{8pt}
\begin{tabular}{ll}
\toprule
\texttt{P0001\_9c030609} & \texttt{P0001\_550ea2ac}  \\
\texttt{P0002\_2ea9af5b} & \texttt{P0009\_02511c2f} \\
\texttt{P0011\_b0c895c1}  & \texttt{P0011\_e5cb6a80} \\
\texttt{P0011\_76ea6d47} & \texttt{P0011\_30a5b61d}  \\
\texttt{P0012\_d6272ce1} & \texttt{P0012\_d85e10f6}  \\
\texttt{P0012\_476bae57} & \texttt{P0015\_e3f96b3f} \\
\bottomrule
\end{tabular}
\end{table}

\paragraph{Training and inference.} We fine-tune Wan2.2-Fun-A14B-Control with rank-32 LoRA and learnable attention bias on the HOT3D training split for 112 GPU hours on a single H200, initializing the gates $g_{\ell,h}$ at zero so that training starts from the pretrained attention. At inference time, we run our geometric action compiler on the raw inputs to create the peeled control video and run our dexterous video generator by sampling 50 steps with a guidance scale of 5, which takes $\sim$ 3.5 minutes on an NVIDIA H100 GPU.

\subsection{Evaluation}
\label{sec:supp:eval}

\paragraph{Protocol.} The methods generate different numbers of frames at different rates, so we fix the time window every method covers rather than its sampling. For a test clip starting at $t_0$, each method receives the frame at $t_0$ as its reference and the action over $[t_0, t_0+T]$ with $T=3.07$\,s (92 source frames at 30\,fps), sampled at its native frame count and rendered at its native frame rate; a method generating $N$ frames thus places frame $i$ at $t_0+iT/(N-1)$, and all outputs share their first and last instants. We then map every output and the ground truth to our 49-frame grid by taking the frame nearest to each grid timestamp. We only drop or repeat frames and never interpolate, so no method is scored on frames it did not generate, and we resize all outputs to $832\times480$. Image metrics therefore compare corresponding frames, and hand, object, and camera trajectories of all methods are estimated at the same instants as the ground truth.

\paragraph{Baselines.} We run each baseline with its released checkpoint, inference script, and default settings, converting our actions to its native interface. Where a method accepts text, we pass the caption of that clip. We caption every training and test clip with Qwen3.5-9B, prompting it to describe the general scene and the hand motion of each hand. EgoHOI receives MANO mesh renders, per-frame camera extrinsics, and the first-frame appearance of the manipulated object, and generates 81 frames. EgoSIM receives its point-cloud prior lifted from $\mathbf{I}_0$, a 2D hand-skeleton video, and a hand-inpainted $\mathbf{I}_0$ from its own editing step, and generates 61 frames. We follow their exact recipe to preprocess our evaluation datasets. Wan-Move receives 2D tracks of object surface points projected through the ground-truth poses, and generates 81 frames. 3DTrajMaster receives the 6-DoF pose sequence with an object description, and generates 49 frames. Only Fun-Ctrl* is trained, on our split with the same schedule as HOIWorld. We note that the Fun-Ctrl* baseline partially replicates our Geometric Action Compiler, but with actions overlaid as plain depth instead of peeled depth, while omitting action-biased attention.

\paragraph{Metrics.} We run WiLoR \cite{Potamias2024WiLoRE3}, FoundationPose \cite{Wen2023FoundationPoseU6}, and MapAnything \cite{Keetha2025MapAnythingUF} from their official repositories with default settings, identically on generated and ground-truth videos. For PA-MPJPE and PCK@20\,mm, we run WiLoR on both videos, Procrustes-align the joints on each generated frame to the WiLoR estimate on the corresponding ground-truth frame, and exclude frames without a detectable hand on either side. We initialize FoundationPose with the object's CAD model, Depth-Anything-V2 metric depth (scaled using the ground-truth first-frame object mask and pose), and the ground-truth first-frame object mask, and report ADD-S and rotation error over tracked frames together with the detection rate. We align camera trajectories with Sim(3) before computing ATE, RRE, and rotational RPE. We compute FID with Inception-v3 features over aligned frames, and FVD with I3D features (one embedding per clip).

\section{Ablations}
\label{sec:supp:ablation}

\paragraph{Scene reconstruction error.} We measure how accurately the first-frame scene layer recovers the geometry against which the action is rendered. On HOT3D, we compare the depth of reconstructed scene point cloud $\mathcal{S}_0$ with the rendered ground-truth depth of the static objects visible in $\mathbf{I}_0$, which the scale estimate does not use. In \cref{tab:supp:scale}, we report this error for Pi3X before and after rescaling, and for a privileged reconstruction from all frames of the clip. Rescaling reduces the median error from 36.5 to 3.8 cm, and the remaining gap to the all-frames reconstruction reflects the depth ambiguity of a single view.

\begin{table}[h]
\centering
\small
\caption{\textbf{Scene reconstruction error} (HOT3D), at pixels of $\mathbf{I}_0$ covered by static objects. The all-frames reconstruction uses 24 frames sampled at equal intervals from the 81-frame clip.}
\label{tab:supp:scale}
\begin{tabular}{@{}lcc@{}}
\toprule
\textbf{Scene layer} & \textbf{AbsRel}$\downarrow$ & \textbf{Median err. (cm)}$\downarrow$ \\
\midrule
First frame, unscaled & 0.992 & 36.5 \\
First frame, rescaled (ours) & 0.179 & 3.8 \\
All frames, rescaled (privileged) & 0.085 & 1.2 \\
\bottomrule
\end{tabular}
\end{table}

\paragraph{Effect of the scene layer.} We carry this comparison to generation by conditioning HOIWorld on no scene layer, on our first-frame layer, and on the privileged all-frames reconstruction, an oracle that sees scene geometry from the whole clip and is unavailable at deployment. We train all three variants for hand control for 5k steps with the same peeled layer and action-biased attention, so that they differ only in the scene layer, and evaluate them on 250 HOT3D test clips (\cref{tab:supp:scene}). Removing the scene layer degrades every metric, including hand accuracy, as the model loses both the ego-motion signal and the surfaces that place the hands in the scene. Our first-frame layer closes nearly all of the gap between no scene layer and the oracle on hand accuracy, and about three quarters of it on PSNR and RRE. The remaining gap lies mainly in camera motion, where the oracle reduces RRE from 26.2 to 20.5$^\circ$, which we attribute to its more accurate geometry and its coverage of regions that the camera reveals after the first frame.

\begin{table}[t]
\centering
\small
\caption{\textbf{Effect of the scene layer} (HOT3D, hand control, 250 test clips). All variants are trained for 5k steps. MPJPE in mm, RRE in degrees.}
\label{tab:supp:scene}
\resizebox{\columnwidth}{!}{%
\begin{tabular}{@{}l@{\hspace{3pt}}ccccc@{}}
\toprule
\textbf{Scene layer} & \textbf{PSNR}$\uparrow$ & \textbf{LPIPS}$\downarrow$ & \textbf{MPJPE}$\downarrow$ & \textbf{PCK@20}$\uparrow$ & \textbf{RRE}$\downarrow$ \\
\midrule
None & 17.72 & 0.336 & 28.15 & 62.8 & 42.8 \\
First frame (ours) & 20.13 & 0.265 & 18.23 & 79.7 & 26.2  \\
All frames (privileged) & 20.79 & 0.222 & 18.04 & 80.2 & 20.5 \\
\bottomrule
\end{tabular}
}
\end{table}

\paragraph{Peel encoding.} The peeled layer differs from plain depth in two ways: it retains surfaces hidden from the camera, and it draws the action in color, separating it from the grayscale scene. \cref{tab:supp:peel} separates the two. First, we truncate the peel of our full model to its first $K\in\{1,5,12\}$ hits at inference, where $K{=}1$ keeps only the visible surface in color. Truncating to the visible surface raises MPJPE from 14.18 to 24.35 and RotErr from 16.55 to 28.98$^\circ$, showing that the model relies on hidden surfaces. $K{=}5$ recovers most of the hand accuracy but less of the object accuracy (ADD-S of 3.69 vs.\ 2.66\,cm), as rays through a grasped object cross more surfaces than rays through a hand alone. Since truncation also presents inputs unseen in training, we further retrain three alternative encodings together with ours under a 5k-step schedule: grayscale depth of the visible surface, which drops both hidden surfaces and color; a visible-only peel ($K{=}1$), which keeps color but drops hidden surfaces; and layered channels, which store the depths of the first, second, and last hit in the three channels instead of compositing all hits with Eq. 2 of our main paper. Color accounts for most of the gain in hand accuracy, reducing MPJPE from 41.44 to 22.35, while hidden surfaces matter most for object rotation, reducing RotErr from 26.56 to 20.12$^\circ$. Layered channels (1st, 2nd and last hits) fall between the visible-only peel and ours on every metric, so compositing all hits conveys more than a fixed subset of layers. We cap the peel at 12 hits, since only 0.08\% of training rays intersect more than 12 surfaces.

\begin{table}[t]
\centering
\small
\caption{\textbf{Peel encoding} (HOT3D, 250 test clips). Top: our full model with the peel truncated to $K$ hits at inference. Bottom: encodings retrained for 5k steps. MPJPE in mm, ADD-S in cm, RotErr in degrees.}
\label{tab:supp:peel}
\resizebox{\columnwidth}{!}{%
\begin{tabular}{@{}l@{\hspace{4pt}}ccccc@{}}
\toprule
& \multicolumn{3}{c}{\textbf{Hand control}} & \multicolumn{2}{c}{\textbf{Object control}} \\
\cmidrule(lr){2-4} \cmidrule(l){5-6}
\textbf{Encoding} & \textbf{MPJPE}$\downarrow$ & \textbf{PCK@20}$\uparrow$ & \textbf{LPIPS}$\downarrow$ & \textbf{ADD-S}$\downarrow$ & \textbf{RotErr}$\downarrow$ \\
\midrule
\multicolumn{6}{@{}l}{\textit{Full model}} \\
$K=1$  & 24.35 & 74.8 & 0.282 & 6.04 & 28.98 \\
$K=5$ & 14.88 & 83.3 & 0.264 & 3.69 & 18.85 \\
$K=12$ (ours) & 14.18 & 83.9 & 0.260 & 2.66 & 16.55 \\
\midrule
\multicolumn{6}{@{}l}{\textit{Retrained, 5k steps}} \\
Depth only & 41.44 & 59.6 & 0.365 & 7.92 & 30.00 \\
$K=1$ peel & 22.35 & 75.1 & 0.278 & 6.90 & 26.56 \\
Layered (1/2/last) & 19.91 &  78.4 & 0.269 & 6.33 & 21.80 \\
$K=12$ (ours) & 18.23 & 79.7 & 0.265 & 5.98 & 20.12 \\
\bottomrule
\end{tabular}
}
\end{table}

\pgfplotstableread{
id win tie loss sig
0  72.2  5.6  22.2  1
1  100   0    0     1
2  83.3  5.6  11.1  1
}\studyHandHandmot
\pgfplotstableread{
id win tie loss sig
0  70.7  21.2  8.1   1
1  94.4  5.6   0     1
2  77.8  11.1  11.1  1
}\studyHandPlaus
\pgfplotstableread{
id win tie loss sig
0  72.2  16.7  11.1  1
1  83.3  16.7  0     1
2  55.6  33.3  11.1  0
}\studyHandConsist
\pgfplotstableread{
id win tie loss sig
0  94.7  0     5.3   1
1  94.4  0     5.6   1
2  78.4  15.1  6.5   1
}\studyObjObjmot
\pgfplotstableread{
id win tie loss sig
0  73.7  10.5  15.8  1
1  88.9  11.1  0     1
2  73.2  19.3  7.5   1
}\studyObjPlaus
\pgfplotstableread{
id win tie loss sig
0  68.4  31.6  0     1
1  77.8  22.2  0     1
2  56.8  37.9  5.3   0
}\studyObjConsist

\definecolor{studyWin}{HTML}{6C5CE7} 
\definecolor{studyTie}{HTML}{D5D9DF}
\definecolor{studyLoss}{HTML}{8A919B}
\definecolor{studyInk}{HTML}{1F2329}

\newcommand{\studyval}{\pgfkeys{/pgf/fpu=true}\pgfmathparse{round(\pgfplotspointmeta)}%
  \pgfmathfloattoint{\pgfmathresult}\edef\studyV{\pgfmathresult}}
\newcommand{\studybars}[1]{%

  \addplot[fill=studyWin, point meta=explicit,
    nodes near coords={\studyval\edef\studyS{\the\numexpr\studyV/1000\relax}%
      \edef\studyV{\the\numexpr\studyV-1000*\studyS\relax}%
      \ifnum\studyV<15 \else\studyV\ifnum\studyS=1 $^\dagger$\fi\fi},
    every node near coord/.append style={anchor=center, text=white,
      font=\scriptsize\bfseries}]
    table[x=win, y=id, meta expr=\thisrow{win}+1000*\thisrow{sig}] {#1};
  \addplot[fill=studyTie] table[x=tie, y=id] {#1};
  \addplot[fill=studyLoss, point meta=explicit,
    nodes near coords={\studyval\ifnum\studyV<10 \else\studyV\fi},
    every node near coord/.append style={anchor=center, text=studyInk,
      font=\scriptsize}]
    table[x=loss, y=id, meta=loss] {#1};
}

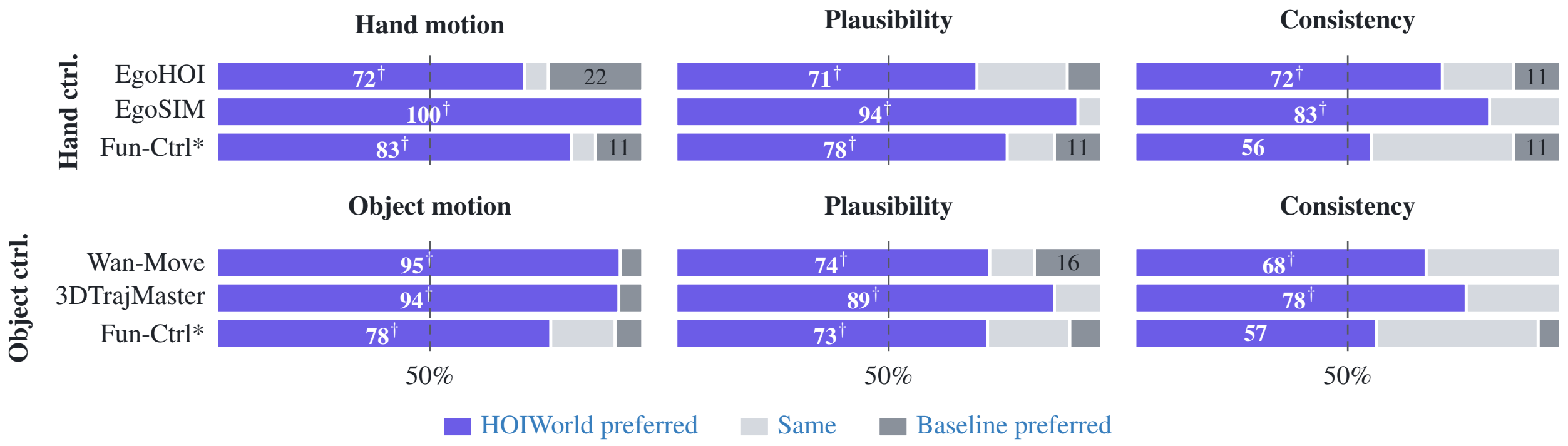
\begin{figure*}[t]
\centering
\begin{tikzpicture}
\begin{groupplot}[
  group style={group size=3 by 2, horizontal sep=10pt, vertical sep=22pt,
               yticklabels at=edge left, xticklabels at=edge bottom},
  scale only axis, width=128pt, height=34pt,
  xbar stacked, /pgf/bar width=9pt, y dir=reverse,
  xmin=0, xmax=100, ymin=-0.6, ymax=2.6, ytick={0,1,2},
  xtick={50}, xticklabels={50\%}, xmajorgrids,
  major grid style={densely dashed, studyInk!75, line width=0.5pt},
  axis on top,
  axis line style={draw=none}, tick style={draw=none},
  every axis plot/.append style={draw=white, line width=0.9pt},
  tick label style={font=\footnotesize, text=studyInk},
  title style={font=\footnotesize\bfseries, text=studyInk, yshift=-3pt},
  ylabel style={font=\footnotesize\bfseries, text=studyInk},
  clip=false,
  legend style={draw=none, font=\footnotesize, legend columns=3,
                /tikz/every even column/.append style={column sep=10pt}},
  legend image code/.code={\fill[#1] (0pt,-3pt) rectangle (9pt,3pt);},
]
  \nextgroupplot[title={Hand motion}, ylabel={Hand ctrl.},
                 yticklabels={EgoHOI, EgoSIM, Fun-Ctrl*},
                 legend to name=studylegend]
    \studybars{\studyHandHandmot}
    \legend{HOIWorld preferred, Same, Baseline preferred}
  \nextgroupplot[title={Plausibility}] \studybars{\studyHandPlaus}
  \nextgroupplot[title={Consistency}]  \studybars{\studyHandConsist}
  \nextgroupplot[title={Object motion}, ylabel={Object ctrl.},
                 yticklabels={Wan-Move, 3DTrajMaster, Fun-Ctrl*}]
    \studybars{\studyObjObjmot}
  \nextgroupplot[title={Plausibility}] \studybars{\studyObjPlaus}
  \nextgroupplot[title={Consistency}]  \studybars{\studyObjConsist}
\end{groupplot}
\end{tikzpicture}\\[1pt]
\ref{studylegend}
\caption{\textbf{User study.} Pairwise comparisons of HOIWorld against each baseline, with the real video shown for reference; bars give the share of votes preferring HOIWorld (purple), rating both the same, or preferring the baseline. The motion column asks which video's controlled motion (hands under hand control, object under object control) is closer to the real video. $^\dagger$\,HOIWorld's win rate among decisive votes is significantly above 50\% (95\% bootstrap CI over participants).}
\label{fig:supp:study}
\end{figure*}

\paragraph{Object control by difficulty.} We split the HOT3D object-control test clips at the median of total object rotation and of hand occlusion, measured as the per-clip mean fraction of object mesh pixels whose first hit is a hand, and report Fun-Ctrl* and \method on each half in \cref{tab:supp:strat}. \method outperforms Fun-Ctrl* on every half, and its advantage in ADD-S grows with difficulty, from 1.6 to 4.7\,cm across rotation and from 1.0 to 5.3\,cm across occlusion. Occlusion separates the two most: Fun-Ctrl* triples its ADD-S from the low to the high half (2.92 to 8.76\,cm), while \method degrades less than twofold (1.88 to 3.44\,cm), consistent with the peeled layer retaining the object surfaces that the hands hide from plain depth. The RotErr gap widens with rotation, from 7.8 to 12.5$^\circ$, but remains about 10$^\circ$ across occlusion levels. Fun-Ctrl* also loses more detections under occlusion (95.2 to 85.2\%), so its errors on the high half exclude the frames where the object is lost and, if anything, understate the gap.

\begin{table}[t]
\centering
\small
\caption{\textbf{Object control by difficulty} (HOT3D). Clips are split at the median total rotation (89.6$^\circ$) and median hand occlusion (4.9\%). Gap: improvement of \method over Fun-Ctrl*.}
\label{tab:supp:strat}
\begin{tabular}{@{}lcccc@{}}
\toprule
& \multicolumn{2}{c}{\textbf{Rotation}} & \multicolumn{2}{c}{\textbf{Occlusion}} \\
\cmidrule(lr){2-3} \cmidrule(l){4-5}
& \textbf{Low} & \textbf{High} & \textbf{Low} & \textbf{High} \\
\midrule
\multicolumn{5}{@{}l}{\textit{ADD-S (cm)} $\downarrow$} \\
Fun-Ctrl* & 3.76 & 7.90 & 2.92 & 8.76 \\
\method & 2.12 & 3.20 & 1.88 & 3.44 \\
\textit{Gap} & \textit{1.64} & \textit{4.70} & \textit{1.04} & \textit{5.32} \\
\midrule
\multicolumn{5}{@{}l}{\textit{RotErr ($^\circ$)} $\downarrow$} \\
Fun-Ctrl* & 19.75 & 34.05 & 23.25 & 30.55 \\
\method & 11.98 & 21.52 & 13.52 & 20.01 \\
\textit{Gap} & \textit{7.77} & \textit{12.53} & \textit{9.73} & \textit{10.54} \\
\midrule
\multicolumn{5}{@{}l}{\textit{DetRate (\%)} $\uparrow$} \\
Fun-Ctrl* & 92.10 & 88.38 & 95.15 & 85.18 \\
\method & 96.85 & 93.78 & 99.01 & 90.76 \\
\textit{Gap} & \textit{4.75} & \textit{5.40} & \textit{3.86} & \textit{5.58} \\
\bottomrule
\end{tabular}
\vspace{-1em}
\end{table}

\section{Additional Results}
\label{sec:supp:results}

\paragraph{User study.} Our metrics measure fidelity to the ground truth but not whether an interaction looks physically right, so we ran a blind user study with 18 participants. In our user study, each participant is shown a ground truth video and a pair of generated videos in random order, one of which belongs to our method. Participants are asked to judge 10 hand-control and 10 object-control clips from HOT3D, EgoDex, and OakInk2, drawn at random from a pool of 100 clips, where each trial pairs \method with one baseline: EgoHOI, EgoSIM, or Fun-Ctrl* under hand control, and Wan-Move, 3DTrajMaster, or Fun-Ctrl* under object control. For each trial, participants are asked to choose the better video, or `same', on three questions: (1) \emph{motion fidelity}, whether the controlled motion, \ie the hands under hand control and the object under object control, is closer to the real video; (2) \emph{physical plausibility}, whether hands, objects, and scene interact plausibly, without interpenetration, unsupported motion, or floating objects; and (3) \emph{temporal consistency}, whether hands and objects keep their shape and identity without flicker. We report the share of votes for each outcome in \cref{fig:supp:study} and test whether the win rate of \method among decisive votes exceeds 50\% with a 95\% bootstrap confidence interval over participants.

As shown in \cref{fig:supp:study}, participants prefer \method over every baseline on every question, where \method wins most clearly on motion fidelity, taking 72--100\% of hand-control and 78--95\% of object-control comparisons, which matches the kinematic metrics of the main paper. \method wins at least 68\% of comparisons against EgoHOI, EgoSIM, Wan-Move, and 3DTrajMaster on every question, and it outperforms the two dexterous world models even on their own training data: EgoHOI trains on HOT3D and may have seen our HOT3D test sequences, yet degrades on EgoDex, while EgoSIM trains on EgoDex, where \method runs zero-shot. Participants never prefer EgoSIM over \method, and prefer 3DTrajMaster, a state-of-the-art 6-DoF object trajectory-conditioned video generation method, in only 6\% of motion-fidelity comparisons. Fun-Ctrl*, which shares our backbone and training data, is the closest competitor. \method still wins significantly on motion fidelity and plausibility in both modes, but temporal consistency draws the most ties (33\% and 38\%), and there the advantage of \method is not significant. We attribute this to the shared backbone, which gives both models similar frame-to-frame stability: the peeled layer and the attention bias change what the video shows, namely how closely it follows the action and how plausibly hands and objects interact, rather than how smoothly it evolves over time.

\begin{figure}[h]
\centering
\includegraphics[width=\columnwidth]{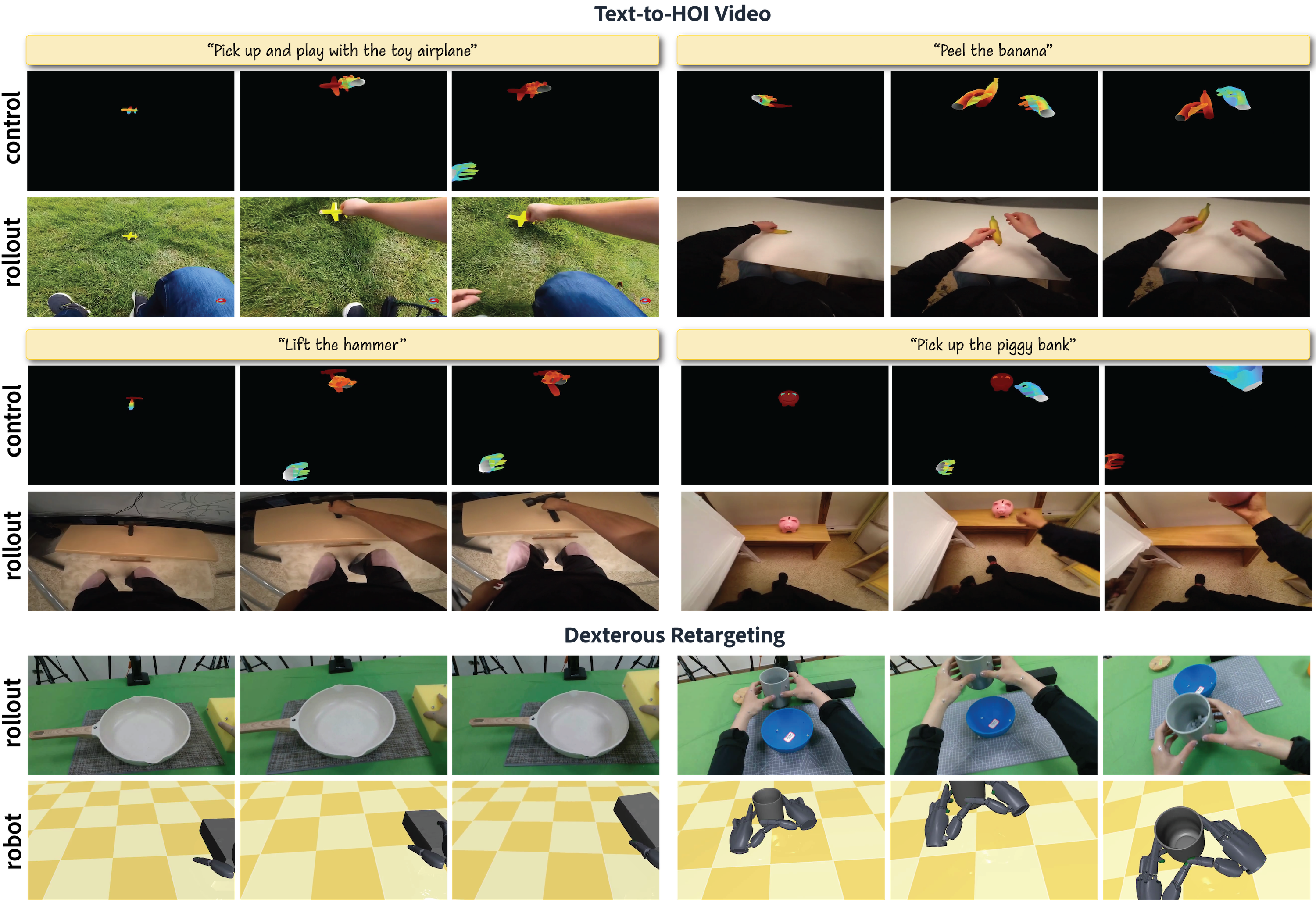}
\caption{\textbf{Applications of \method.} (Top) Sim-to-real generation: we render text-generated Dex2HOI 3D hand-object sequences as peeled control videos without a scene layer and generate videos without a reference frame. (Bottom) Dexterous retargeting: we retarget \method rollouts to the Sharpa Wave hand with Do as I Do and execute them in MuJoCo.}
\label{fig:supp:apps}
\end{figure}

\begin{figure*}[h]
\centering
\includegraphics[width=\textwidth]{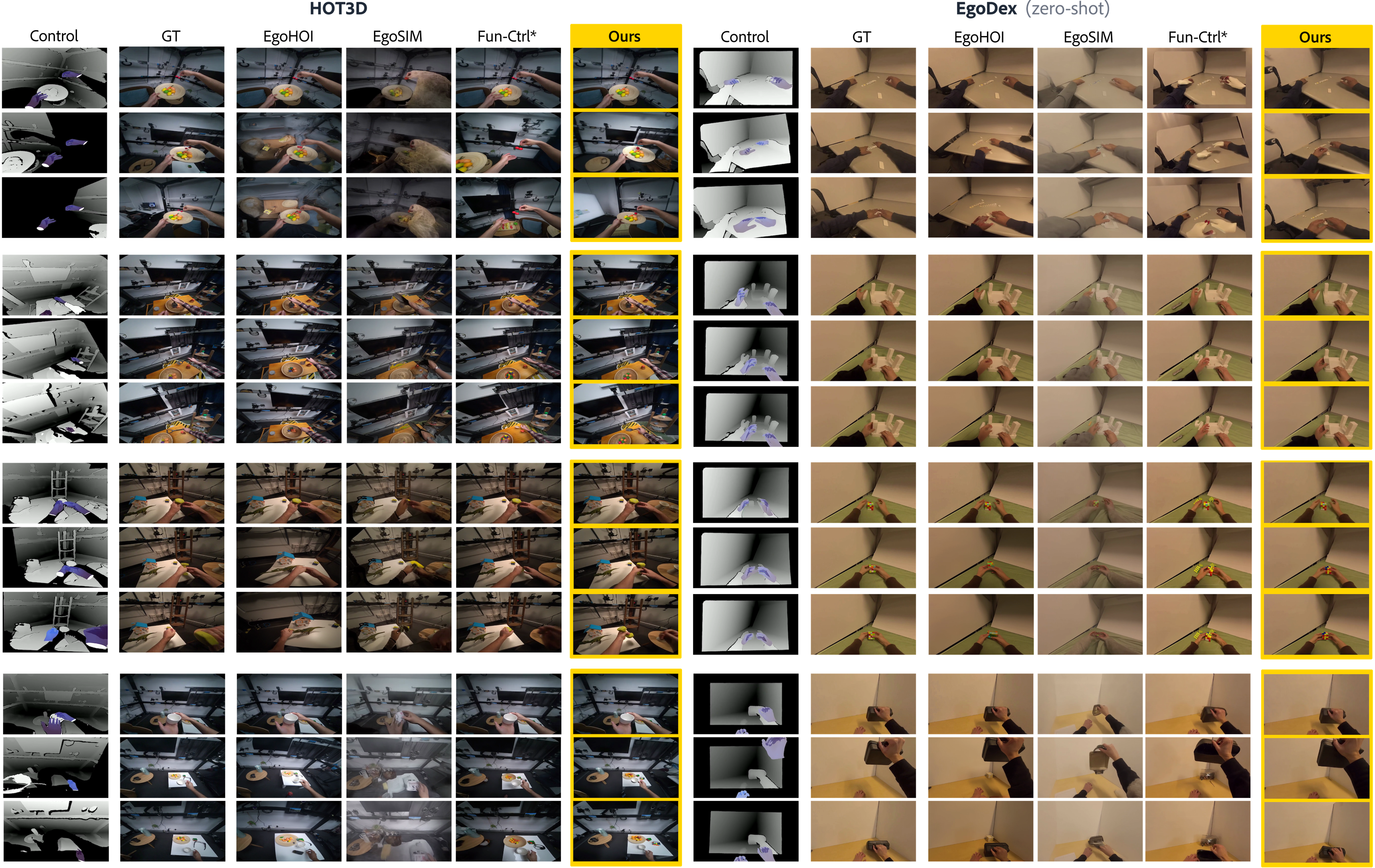}
\caption{\textbf{Hand-controlled generation.} (Left) HOT3D test clips. (Right) Zero-shot EgoDex clips. Columns show the ground truth, our peeled control video, EgoHOI, EgoSIM, Fun-Ctrl*, and \method; only \method takes the peeled control as input, while each baseline receives its native inputs (\cref{sec:supp:eval}). EgoHOI and EgoSIM drift from the reference scene under camera motion, while Fun-Ctrl* is prone to hallucinating objects.}
\label{fig:supp:qual_hand}
\end{figure*}

\begin{figure*}[h]
\centering
\includegraphics[width=\textwidth]{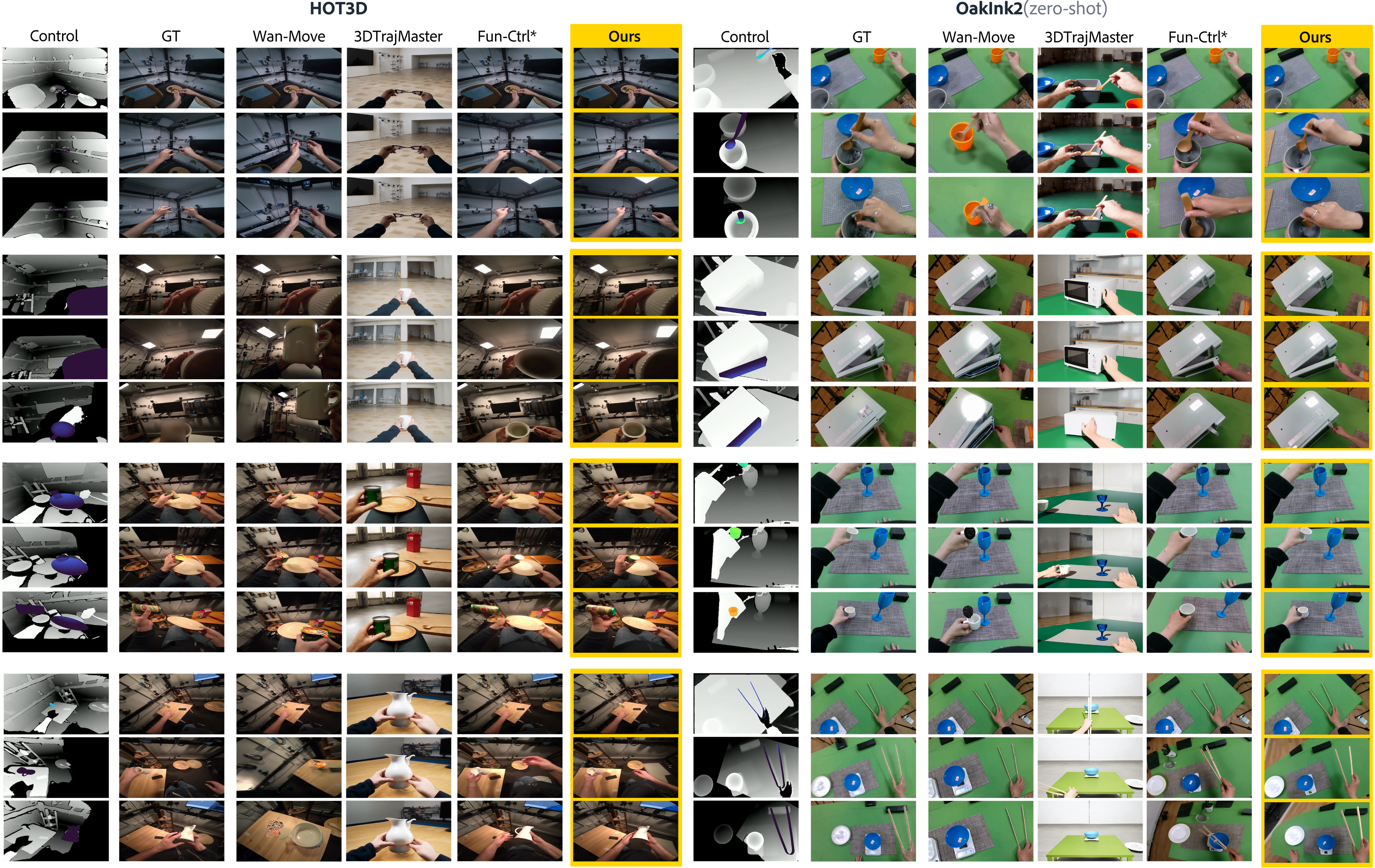}
\caption{\textbf{Object-controlled generation.} (Left) HOT3D test clips. (Right) Zero-shot OakInk2 clips. Columns show the ground truth, our peeled control video, Wan-Move, 3DTrajMaster, Fun-Ctrl*, and \method; only \method takes the peeled control as input, while each baseline receives its native inputs (\cref{sec:supp:eval}). \method generates the hands that move the object along the input trajectory, while Fun-Ctrl* is prone to object jumps and distortions under heavy occlusion.}
\label{fig:supp:qual_obj}
\end{figure*}

\paragraph{Qualitative results.} We show hand-control comparisons on HOT3D and EgoDex in \cref{fig:supp:qual_hand}, object-control comparisons on HOT3D and OakInk2 in \cref{fig:supp:qual_obj}, and sim-to-real generations from Dex2HOI controls without a reference frame and retargeting rollouts executed by the Sharpa Wave hand in MuJoCo in \cref{fig:supp:apps}. Each comparison shows the ground truth, our peeled control video, and the outputs of all methods side by side; only our method takes the peeled control as input, while each baseline receives its native inputs as described in \cref{sec:supp:eval}. As in our evaluation protocol, we resample every method to a common frame rate by nearest timestamp, without frame interpolation. Across these samples, \method follows the input actions more closely than the baselines, while keeping the scene appearance consistent, and reproducing the camera motion. EgoHOI, which may have seen our HOT3D test sequences during training, and EgoSIM both suffer from scene drift and degrade under large camera motion. On EgoDex, which contains little camera motion, \method runs zero-shot and still produces more faithful and more plausible interactions than EgoSIM, even though EgoSIM trains on 400K clips from EgoDex and EgoVid and we train on HOT3D alone. Fun-Ctrl*, the strongest baseline, is prone to hallucinating objects and makes sudden jumps in object pose, particularly when the hand occludes the object. Since it shares our backbone and training data but lacks the peeled layer and the attention bias, these failures show what the two components contribute, consistent with our ablation in the main paper.

\section{Limitations}
\label{sec:supp:fail}
We train \method only on HOT3D, which was captured in a lab environment with a limited set of objects, so our model inherits the resulting biases. Given a reference frame, \method can fail on objects whose shape or appearance differs strongly from those it saw during training. Without a reference frame, as in the sim-to-real setting, it tends to generate indoor scenes. Training on larger and more diverse egocentric data offers a natural way to address both.

\end{document}